\documentclass[lettersize,journal]{IEEEtran}
\usepackage{amsmath,amsfonts}
\usepackage{array}
\usepackage{textcomp}
\usepackage{stfloats}
\usepackage{url}
\usepackage{nccmath}
\usepackage{verbatim}
\usepackage{graphicx}
\usepackage{cite}
\usepackage{textcomp}
\usepackage{overpic}
\usepackage{ragged2e}
\usepackage{gensymb}
\usepackage{amssymb}
\usepackage{mathtools}
\usepackage{xcolor}
\usepackage{float}
\usepackage{multirow}
\usepackage{booktabs}
\usepackage{siunitx}
\usepackage{MnSymbol}
\usepackage[linesnumbered,ruled,vlined]{algorithm2e}
\SetAlCapNameFnt{\footnotesize}
\SetAlCapFnt{\footnotesize}
\usepackage{textgreek}
\usepackage{tcolorbox}
\usepackage[caption=false]{subfig}
\usepackage[pdfencoding=auto, psdextra]{hyperref}

\usepackage{tikz}
\newcommand{\rev}[1]{\textcolor{black}{#1}}
\newcommand{\revtwo}[1]{\textcolor{black}{#1}}
\makeatletter
\renewcommand*{\@algocf@post@ruled}{}
\makeatother

\makeatletter
\def\ps@IEEEtitlepagestyle{%
	\def\@oddfoot{\mycopyrightnotice}%
	\def\@evenfoot{}%
}
\def\mycopyrightnotice{%
	{\footnotesize SUBMITTED FOR PEER REVIEW\hfill}
	\gdef\mycopyrightnotice{}
}

\begin{document}
	
	\title{Tactile-based Exploration, Mapping and Navigation \\ with Collision-Resilient Aerial Vehicles}
	
	\author{Karishma Patnaik, Aravind Adhith Pandian Saravanakumaran and Wenlong Zhang$^{*}$
		\thanks{This material is based upon work supported by the National Science Foundation under Grant No. 2331781.}
		\thanks{The authors are with School of Manufacturing Systems and Networks, Ira A. Fulton Schools of Engineering, Arizona State University, Mesa, AZ, 85212, USA. 
			Email: {\tt\small $\{$kpatnaik, apandia1, wenlong.zhang$\}$@asu.edu}.}%
		\thanks{$^{*}$Address all correspondence to this author.}%
	}
	
	\graphicspath{{figures/}}
	\maketitle

	\begin{abstract}
		This article introduces XPLORER, a passive deformable quadrotor UAV with a spring-augmented chassis and proprioceptive state awareness, designed to endure collisions and maintain smooth contact with the environment. A fast-converging external force estimation algorithm for XPLORER is designed to leverage onboard sensors and proprioceptive data for contact detection. Using this force information, four motion primitives are proposed, including three novel tactile-based primitives—tactile-traversal, tactile-turning, and ricocheting—to aid XPLORER in navigating unknown environments. These primitives are synthesized autonomously in real time to enable efficient exploration and navigation by leveraging collisions and contacts. Experimental results demonstrate the effectiveness of our approach, highlighting the potential of \textit{passive} deformable UAVs for contact-rich real-world tasks such as non-destructive inspection, surveillance and mapping, and pursuit/evasion.
	\end{abstract}
	\vspace{-0.2in}

	\section{Introduction}
	
	\revtwo{Conventional multirotor unmanned aerial vehicles (UAVs), often rely on external attachments or complex actuation mechanisms for tasks such as grasping, perching, or pushing/pulling \cite{RL18,HZ+05,MS+21,KC+13,SH+15,B+24}. While reconfigurable UAVs offer improved flight efficiency over rigid designs, they still rely on prior knowledge of the environment \cite{K16,MF16,TP23,PW23}. These platforms often struggle to operate autonomously in cluttered, unstructured environments—such as during search and rescue, disaster response and pursuit/evasion—where contact with unknown surfaces is inevitable. Limited compliance makes such UAVs vulnerable to unpredictable behaviors upon collision, such as bouncing, instability, or complete mission failure. This severely constrains their ability to safely navigate confined spaces 
		or adapt to complex surface geometries \cite{PW21}.} 

\revtwo{In contrast, passive deformable UAVs adapt their morphology naturally in response to contact forces, using springs, origami structures, or soft bodies to absorb impact energy and retain stability \cite{PM+20,LK21,PM+21,PK+23}. This article advances the scope of passive deformable UAVs by introducing new tactile-based planning and control algorithms for reliable performance in demanding real-world scenarios.}
\begin{figure}[t]
	\vspace{-0.1in}
	\centering
	\subfloat{\includegraphics[trim = 0cm 1.5cm 0cm 0cm, clip, width = 0.48\textwidth]{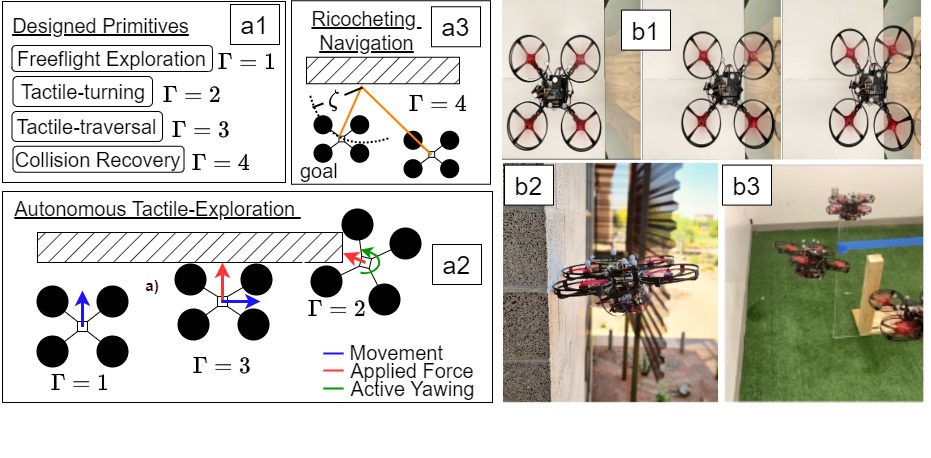}}
	\vspace{-0.1in}
	\caption{An overview of the article. (a1) The four primitives ($\Gamma$) proposed to facilitate autonomous missions with deformable UAVs. (a2)-(a3) Automatic online synthesis of $\Gamma$ for tactile-based exploration and navigation. (b1) Sequence of slow-motion snapshots highlighting the chassis deformation upon contact. (b2) XPLORER enables tactile-based mapping of buildings using (a2). (b3) Ricocheting for agile maneuvering using (a3).  
	}
	\label{fig:first_pic}
	\vspace{-0.2in}
\end{figure}
\vspace{-0.1in}
\subsection{Related Literature}

\subsubsection{External Wrench Estimation}
Augmenting a UAV with force-torque sensors is the most common approach to detect external wrench, but it increases the total weight of the UAV and adversely affects flight time \cite{SP+20,OH+18}.
Alternatively, researchers have developed algorithms based on momentum-based \cite{RC+14,RC+15} and acceleration-based  \cite{YS+14,RM+19} wrench estimators which rely on the available onboard sensors only such as inertial measurement units (IMUs). Kalman filtering approaches and hybrid methods, fusing momentum and acceleration-based wrenches, were proposed in \cite{MS16,TOH17} for conventional UAVs. \textcolor{black}{For this work, we aim to exploit \textit{proprioceptive} information about the vehicle's morphology to improve the force estimate and identify the colliding arms.}

\subsubsection{Static Force Application and Tactile-based Inspection}
Using the external wrench estimate, UAVs are now capable of performing tactile-based tasks such as non-destructive testing, vibration analysis, and leak identification which require a desired wrench to be applied onto a surface in an attempt to reduce safety risks to human workers \cite{PB+19,HZ22,AD16,KB+20,TC+19}.
\textcolor{black}{In this context, this work employs \textit{passive} deformable quadrotors to establish smooth contacts and simplify the contact laws}  \cite{B99}. 

\subsubsection{Tactile-based Navigation}
Collision-resilient designs hold a strong potential for a paradigm shift for performing tactile-based navigation \cite{ZB10,SB09,HB10,SM11,SA14}. A motion planner for tactile-based navigation was proposed in \cite{KM19} consisting of two modes, sliding and flying cartwheel. Authors of \cite{DN21} utilized vision systems to navigate manhole-sized tubes along with flaps to sense contacts. 
In \cite{ZM21}, collisions were exploited for path planning using sampling-based methods. However, all the works assume prior complete knowledge of the environment. 

\subsubsection{Mapping}
Conventionally, mapping an environment is performed by using cameras and Structure from Motion (SfM) technique \cite{SF16} by recreating a 3D structure from a series of images captured from different angles. 
Research carried out in \cite{CV19} used the combination of cameras, LiDAR, IMU, and encoders to generate a highly precise map of the environment. Another map generation methodology was presented in \cite{TOH17} where collision detection and localization were used to insert obstacle blocks into the point cloud map. However, autonomous planning for performing these missions in unknown low-visible environments where collisions are inevitable remains a challenge. Moreover, previously studied collision-based mapping \cite{ML20,BK13} are error-prone particularly when mapping curved surfaces, making continuous contact-based mapping a desirable approach, as shown in this paper.

\vspace{-0.1in}
\subsection{Contributions of Present Work}
\revtwo{We introduce a novel set of tactile motion primitives for \textit{deformable quadrotors}, enabling real-time contact-based path adaptation and reducing the need for meticulous planning in unstructured environments.} The main contributions are:
\begin{enumerate}
	\item A fast-converging external force estimation algorithm for deformable quadrotors, leveraging proprioceptive morphology and onboard accelerometer data.
	\item \revtwo{Two novel tactile motion primitives for exploration, \textit{Tactile-traversal, Tactile-turning}, that adaptively modify the UAV’s references based on real-time contact forces. 
		\item A novel 2D \textit{Ricocheting} primitive enabling efficient braking and minimum-time maneuvers by utilizing collision energy dissipation. }
\end{enumerate}

Figure \ref{fig:first_pic} and Supplementary Video (Part1) showcase our compliant and passive deformable quadrotor, XPLORER, engaged in various contact-rich tasks. \revtwo{Additionally, Supplementary Material \href{https://arxiv.org/pdf/2305.17217}{Suppl.} includes a literature table detailing the current state-of-the-art from the previous section.}

\begin{table}
	\centering
	\caption{}\label{table:nomen}
	\vspace{-0.1in}
	\renewcommand{\arraystretch}{1.1}
	\begin{tabular}{p{3.2cm}p{5cm}}
		\hline 
		\textbf{Symbol} & \textbf{Definition}\\
		\hline
		$_w\mathcal{F} = \{ \boldsymbol{e_1, ~e_2,~ \boldsymbol{e_3}} \}$ & inertial frame  \\
		$_b\mathcal{F} = \{\boldsymbol{\boldsymbol{b_1},~\boldsymbol{b_2},~b_3} \}$ & body fixed frame \\
		$_{a_i}\mathcal{F} = \{\boldsymbol{a_1^i,~a_2^i}\}$ & $i^{th}$ arm frame with $a_j^i~ \forall j = 1,2$ denoting the basis vectors for the $i^{th}$ frame\\
		$\boldsymbol{x} = [x~ y~ z]^T \in \mathbb{R}^3$ & 3D position of UAV in $_w\mathcal{F}$ \\
		$\boldsymbol{v} = [\dot{x} ~\dot{y} ~\dot{z}]^T \in \mathbb{R}^3 $ & 3D translational velocity of UAV in $_w\mathcal{F}$\\
		$m \in \mathbb{R},g \in \mathbb{R}, $ & mass and gravitational acceleration resp.\\
		$\boldsymbol{R} \in \mathbb{R}^{3\times3}$ & rotation matrix \\
		$\psi \in \mathbb{R}, \dot{\psi} \in \mathbb{R}$ & yaw and yawrate respectively \\
		$\boldsymbol{\hat{\delta}_f}_i \in \mathbb{R}^3$ & estimated force on the $i^{th}$ arm in $_w\mathcal{F}$\\
		$\boldsymbol{\hat{\delta}_f}_{CoM} \in \mathbb{R}^3$ & CoM-based estimated force in $_w\mathcal{F}$\\
		$\boldsymbol{\hat{\delta}_f} \in \mathbb{R}^3$ & net fused estimated force in $_w\mathcal{F}$\\
		$\hat{\delta}_{\tau}^{\boldsymbol{e_3}} \in \mathbb{R}$ & estimated \textit{yaw} torque in $_w\mathcal{F}$\\
		$(\cdot)^{(\bullet)}$ & specifies quantity ($\cdot$) in $(\bullet)$ frame/direction \\
		$\boldsymbol{\hat{\delta}_f^b} \in \mathbb{R}^3 = [\hat{\delta}_f^{\boldsymbol{\boldsymbol{b_1}}}~\hat{\delta}_f^{\boldsymbol{\boldsymbol{b_2}}}~\hat{\delta}_f^{b_3}]^T $ & denotes $\boldsymbol{\hat{\delta}_f}$ in $_b\mathcal{F}$\\
		$\boldsymbol{C}_n \in \mathbb{R}^4$ & contact normal to the obstacle \\
		$\lambda  \in \{ +X, +Y, -X, -Y\}$ & move direction of the vehicle \\
		$\Gamma \in \mathbb{R} $ & indicates the primitive currently engaged \\
		$d_{step} \in \mathbb{R} $ & distance to fly forward in $\Gamma = 1$\\
		$\dot{\psi_0} \in \mathbb{R} $ & yaw rate threshold to set $ \Gamma_{i, i = 1,3} \rightarrow \Gamma = 2$\\
		${\delta}_0 \in \mathbb{R} $ & force threshold to set $\Gamma = 1 \rightarrow \Gamma = 3$ \\
		${{{\delta_{\psi_{0}}}}} \in \mathbb{R} $ & force threshold to set $\Gamma = 2 \rightarrow \Gamma = 3$ \\
		$\boldsymbol{\delta_{f_{des}}} \in \mathbb{R} $ & force for controlled sliding in $\Gamma = 3$\\
		$\mathcal{W} \in \mathbb{R} $ & yaw rate for controlled turning in $\Gamma = 2$\\
		\hline
	\end{tabular} 
	\vspace{-0.2in}
\end{table}

\section{Design and Low-level Control of XPLORER}\label{sec:system}
In this work, we extend our previous work \cite{PM+20} to develop the deformable quadrotor, XPLORER. Based on a similar concept, XPLORER (Fig. \ref{fig:first_pic}) features four deformable arms as part of its morphing chassis, with all four motors aligned in a single plane. This design was chosen for two key reasons. First, it lowers the center of mass (CoM) compared to our previous design (where the arms were in different planes), reducing the risk of toppling upon contact by shortening the lever arm length during impact. Second, the redesigned propeller guards prevent the UAV from getting stuck in corner cases during exploration, allowing up to 30$\degree$ of free rotation. Additionally, the current version uses stiffer springs, limiting collision-induced deformation to less than 30$\degree$. This enables the UAV to exert forces of approximately 1N on the environment without significant pitching, unlike rigid UAVs, which experience at least 5$\degree$ of pitch under similar conditions (e.g., a 1.3kg UAV with a thrust-to-weight ratio of 2.82).

\begin{figure*}
	\centering
	\includegraphics[trim = 0.5cm 0cm 7.35cm 0cm, clip,width = 0.99\textwidth]{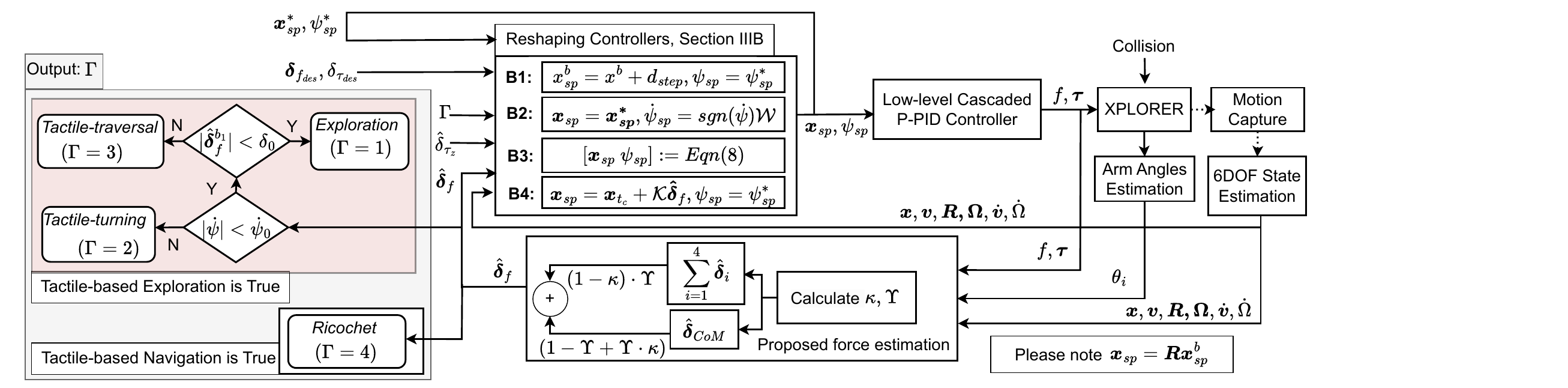}
	\vspace{-0.1in}
	\caption{\revtwo{Block diagram for autonomous tactile exploration and navigation with XPLORER. The external contacts/collisions, characterized in the proposed force estimation block, decide the state of exploration ($\Gamma$) as shown by the flow-chart example (in red). Four reference reshaping controllers are then synthesized to execute the selected $\Gamma$, requiring additional inputs of the current reference, $[\boldsymbol{x}_{sp}^*~\psi_{sp}^*]^T$, current wrench $[\boldsymbol{\hat{\delta}_f}~\hat{\delta}_{\tau_z}]^T$ and desired wrench $[\boldsymbol{\delta_f}_{des}~\delta_{\tau_{des}}]^T$.}}
	\label{fig:cbd}
	\vspace{-0.2in}
\end{figure*}

The complete control block diagram is shown in Fig. \ref{fig:cbd} with the notations defined in Table \ref{table:nomen}.
The dynamics and other symbols adhere to literature and are described in  \href{https://arxiv.org/pdf/2305.17217}{Suppl.} \revtwo{A} with the terms $ \boldsymbol{\delta_{f}} \in \mathbb{R}^3$ and $ \boldsymbol{\delta_{\tau}} \in \mathbb{R}^3$ denoting the lumped external forces and torques respectively applied on the system and $f\in \mathbb{R}, \boldsymbol{\tau} \in \mathbb{R}^3$ being the control thrust and torques respectively. 
The low-level tracking controller is a cascaded P-PID controller that regulates position and attitude errors.
\revtwo{Our previous work has successfully demonstrated the robustness of this structure to small inertia errors (arising from short-term ($\approx10\degree$) arm angle deviations upon collisions) when paired with a deformable chassis \cite{PM+21,PK+23}. Hence, we continue to leverage this framework for precise tracking.}
\section{External Force Estimation and Reaction}\label{sec:wrench_int}
\revtwo{To aid in demonstrating the advantages of deformable chassis, the force estimator should: (1) preserve peak forces upon contact and (2) settle quickly after impact. Towards this, we present the novel force estimator that fuses the proprioceptive state and accelerometer data for faster convergence to true force and isolate the collision location.} 
\vspace{-0.15in}
\subsection{Proposed External Force Estimation Algorithm}\label{sec:force_est}
\subsubsection{Force estimate by the spring action at arm}\label{sec:arm_wrench}
\begin{figure}[t]
	\centering
	\includegraphics[trim = 0.5cm 0.1cm 0.15cm 0cm, clip, width = 0.48\textwidth]{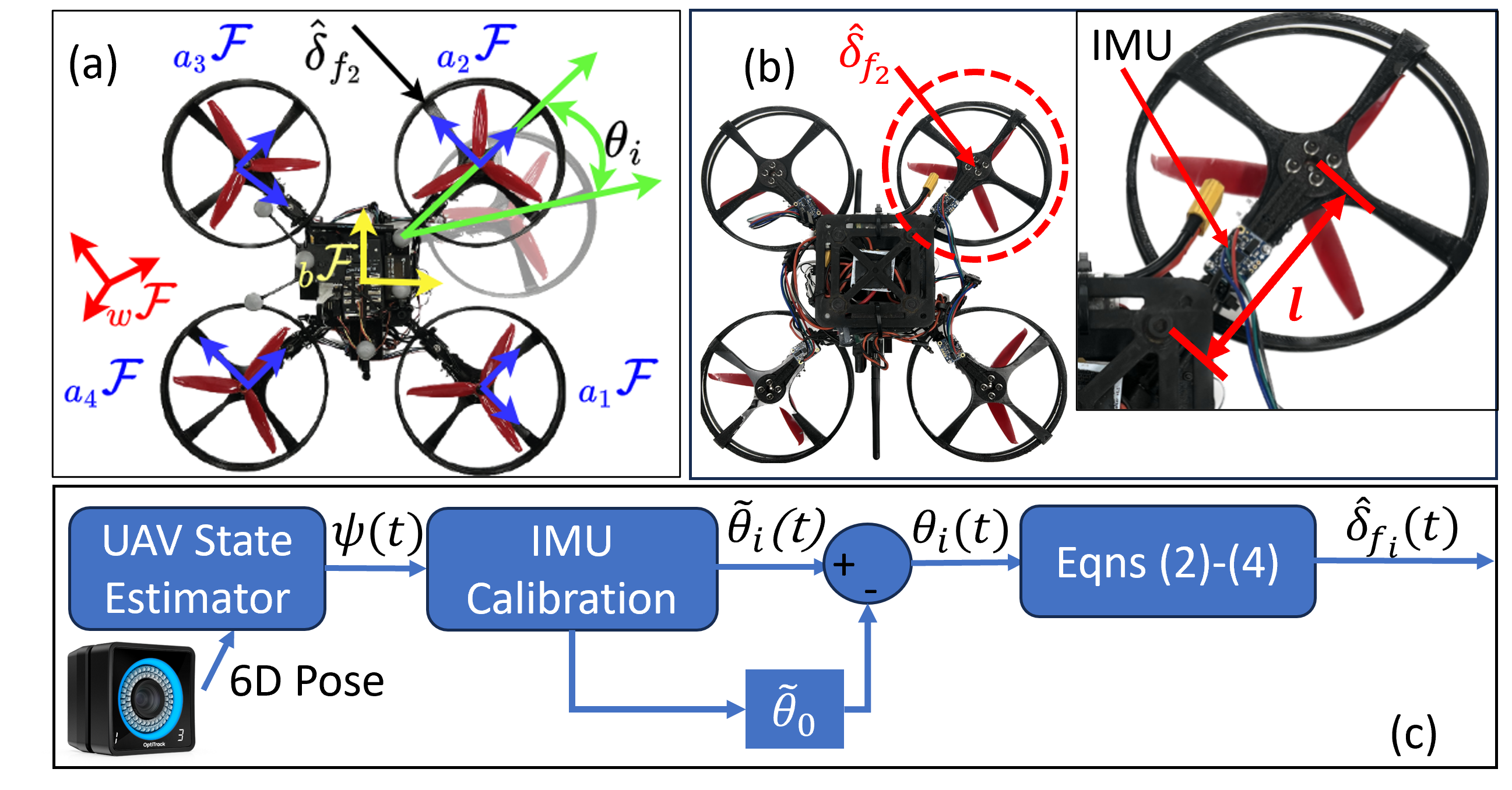}
	\vspace{-0.1in}
	\caption{(a) Frames used to estimate the external force on $i^{th}$ arm, $\boldsymbol{\hat{\delta}_{f_i}}$ (b) A zoomed-in view to show the placement of IMU and the assumed point of application of $\boldsymbol{\hat{\delta}_{f_i}}$ (c) Block diagram for force estimation by spring action.}
	\label{fig:wrench_fbd}
	\vspace{-0.2in}
\end{figure}

\rev{There are four reference frames $_{a_i}\mathcal{F}, i = 1..4$, for each arm as shown in Fig. \ref{fig:wrench_fbd}. Now, considering the $i^{th}$ arm and frame, torque estimate ${\delta}_{\tau_i}^{a_i} \in \mathbb{R}$ can be computed by directly using the angular acceleration and rearranging the arm dynamics equation:
	\begin{equation}
		\mathcal{J}_{zz}\Ddot{\theta}_i + b\dot{\theta}_i + k\theta_i = {\delta}_{\tau_i}^{a_i}  \label{eqn:arm_dynamics}, ~~~   {\delta}_{\tau_i}^{a_i}  = {\delta}_{f_i}^{a_i} l
	\end{equation}
	Here $\mathcal{J}_{zz}$ is the inertia of the arm about the rotation axis, $b$ and $k$ are the damper and spring coefficients respectively and $\theta_i$ is the deflection in the angle for the $i^{th}$ arm. The corresponding external force can be computed from ${\delta}_{\tau_i}^{a_i}  = {\delta}_{f_i}^{a_i} l$ by assuming that the point of application is at the motor center (which is approximately the arm CoM using SolidWorks estimate), at a constant distance $l$ from the spring mounting location as shown by Fig. \ref{fig:wrench_fbd}. Finally, a first-order low-pass filter is applied, since acceleration can be noisy, to estimate $\boldsymbol{{\hat{\delta}}^{a_i}_{f_i}} \in \mathbb{R}^2$, in arm frame: 
	\vspace{-0.1in}
	\begin{equation}
		\begin{aligned}
			\boldsymbol{\dot{\hat{\delta}}^{a_i}_{f_i}} &= [K_I( \delta_{f_i}^{a_i} - \hat{\delta}_{f_i}^{a_i})~0 ]^T
		\end{aligned}
	\end{equation}
	where each element corresponds to the components in the direction of the $i^{th}$ arm's basis and $K_I$ denotes the filter gain.} 
Now the external force on the $i^{th}$ arm in the inertial frame, $\boldsymbol{\hat{\delta}}_{\boldsymbol{f}_i} \in \mathbb{R}^{3}$, can be computed by appropriate rotation (\href{https://arxiv.org/pdf/2305.17217}{Suppl.} B) 
\begin{equation}\label{eqn:arm_wrench}
	\boldsymbol{\hat{\delta}}_{\boldsymbol{f}_{i}} = [^{w}_{a_i}\boldsymbol{R} ~ (\boldsymbol{{\hat{\delta}}_{f_i}^{a_i}}) ~0 ]^T 
\end{equation} 
with $ ^{w}_{a_i}\boldsymbol{R}$ being the rotation from $_{a_i}\mathcal{F}$ frame to $_w\mathcal{F}$ frame. 

\subsubsection{Force estimate from accelerometer at CoM}
Next, following \cite{TOH17}, \rev{force information can also be directly estimated using the vehicle’s acceleration measurements from the inertial measurement unit (IMU) onboard the flight controller.} 
Upon rearranging the translational dynamics of the UAV and using a median low-pass filter, we get the CoM-based estimate as:
\begin{equation}\label{eqn:com_wrench}
	\begin{aligned}
		\hat{\boldsymbol{\delta}}_{\boldsymbol{f}_{CoM}} &= m\boldsymbol{\dot{v}} - m g \boldsymbol{e}_3 + f \boldsymbol{R}\boldsymbol{e}_3 \\
	\end{aligned}
\end{equation}


\subsubsection{Net estimated force on XPLORER}
\rev{The external force obtained from the IMU at CoM has a slower convergence rate than the arm-based estimate. Furthermore, if there is no arm deflection from the external force (e.g. when the force is directly applied at CoM), the estimate based on only arm angle feedback does not converge to the true force, while the estimate from the CoM is correct in this case. In order to make the best use of the two resources and obtain a reliable force estimate, we combine the estimates from both methods. First, define an indicator function to detect contact on the arms
	\vspace{-0.05in}
	\begin{equation}\label{eqn:contact_indicator}
		\Upsilon = \begin{cases}
			1, & \text{if~} \sum_{i=1}^4 |\theta_i| > \theta_{th} \\
			0, & \text{otherwise}
		\end{cases}
\end{equation}}
\hspace{-0.06in}where $\theta_i$ is the deflection angle of each arm, and $\theta_{th}$ is the deflection threshold to detect contact, accounting for backlash. Next, our method for  $\boldsymbol{\hat{\delta}_f}$ fuses $\boldsymbol{\hat{\delta}}_{\boldsymbol{f}_i}$  \ref{eqn:arm_wrench} and (\ref{eqn:com_wrench}) to give a fast and accurate estimate with collision arm isolation:
\vspace{-0.1in}
\begin{equation}\label{eqn:ext_wrench}
	\begin{aligned}
		\boldsymbol{\hat{\delta}_f} &=  \Big(1- \Upsilon + \Upsilon \cdot \boldsymbol{\kappa}\Big) \cdot  \hat{\boldsymbol{\delta}}_{\boldsymbol{f}_{CoM}} + \Big(1-\boldsymbol{\kappa} \Big) \cdot \Upsilon \cdot \sum_{i=1}^4  \boldsymbol{\hat{\delta}}_{\boldsymbol{f}_i} \\
	\end{aligned}
	\vspace{-0.1in}
\end{equation}
Here $\boldsymbol{\kappa} \in \mathbb{R}^3 = \xi{\boldsymbol{\dot{\hat{\delta}}_f}}_{{CoM}}^{|\cdot|} \triangleeq h(\boldsymbol{\dot{\hat{\delta}}_f}_{CoM})$ is an adaptive gain with the tuning parameter $\xi$. $(\bullet)^{|\cdot|}$ denotes the element-wise absolute value of the vector $(\bullet)$. Based on (\ref{eqn:contact_indicator}) and (\ref{eqn:ext_wrench}), one gives higher weightage to $\hat{\boldsymbol{\delta}}_{\boldsymbol{f}_{CoM}}$ during collision since the rate of change for $\hat{\boldsymbol{\delta}}_{\boldsymbol{f}_{CoM}}$ is higher at the collision instant, so $\boldsymbol{\kappa}$ is larger and the CoM-based estimate is followed. On the other hand, when the wrench estimate from the CoM converges slowly, the rate of change is near zero making the term $(1-\boldsymbol{\kappa})$ is larger, so the total wrench converges to the arm-based estimate. \textcolor{black}{The boundedness and convergence properties of the estimator in (\ref{eqn:ext_wrench}), and characteristics of $h$ are discussed in \href{https://arxiv.org/pdf/2305.17217}{Suppl.} C.} Additional details for (\ref{eqn:arm_dynamics}) are in \href{https://arxiv.org/pdf/2305.17217}{Suppl.} D and E. Finally, a momentum-based torque observer \cite{TOH17} is implemented to estimate the external torque $\boldsymbol{\hat{\delta}_\tau} \in \mathbb{R}^3$. However, only the estimated yaw torque, $\hat{\delta}_{\tau}^{\boldsymbol{e_3}} \in \mathbb
R$, is utilized in tactile-based exploration of the following sections. 
\subsection{Reaction Modes and Controllers for Reference Reshaping }\label{sec:int_ctrl}
We now introduce the various reference-generation controllers, which reshape the position, yaw, or both references based on external wrench information, enabling different reaction modes.

For all the subsections that follow, $\boldsymbol{x}_{sp}^* \in \mathbb{R}^3$ and $\psi_{sp}^* \in \mathbb{R}$ are the previous reference values of position and yaw respectively in $_w\mathcal{F}$ which are reshaped into $\boldsymbol{x}_{sp} \in \mathbb{R}^3$ and $\psi_{sp} \in \mathbb{R}$ as the final reference position and yaw, in $_w\mathcal{F}$, for the low-level controller to follow. The complete control block diagram with the interaction controller for XPLORER is shown in Fig. \ref{fig:cbd}. 

\subsubsection{Free-flight with position disturbance rejection}\label{sec:free_flight} 
In this state, at each instant, the position reference in the body frame for any one direction in $x-y-z$ is generated using the current value. For example, if we need to only fly forward in $_b\mathcal{F}$, only $x^b$ will be reshaped, i.e, $x_{sp}^b = x^b + d_{step}$ without $y-z$ or yaw setpoint update. This mode is employed for free-flight exploration and tactile-traversal in Section \ref{sec:tactile_exp}. 

\subsubsection{Continuous yawing}\label{sec:cont_yaw} The UAV \textit{remains} at its current position ($\boldsymbol{x}_{sp} = \boldsymbol{x}$) and continuously turns at a pre-specified constant yaw rate, i.e. $\dot{\psi}_{sp} = sgn(\dot{\psi})\mathcal{W} $ where $\mathcal{W} $ is a user-defined constant, $sgn$ denotes the sign function.

\subsubsection{Static-wrench application}\label{sec:static_wrench} The reference is reshaped to exert a specific wrench on an object and navigate around it as described in Section \ref{sec:tactile_exp}. A position admittance control strategy is employed towards this by modeling the desired response as a virtual second-order dynamics \cite{TOH17,PM+20}:
\begin{equation}\label{eqn:interaction_controller}
	\footnotesize
	\begin{bmatrix}
		m_v \boldsymbol{I}_{3\times3} & \boldsymbol{0}_{3\times1} \\
		\boldsymbol{0}_{1\times3} & \mathcal{I}_{v,z}
	\end{bmatrix}
	\begin{bmatrix}
		\Ddot{\boldsymbol{x}}_{sp} \\
		\Ddot{\psi}_{sp}
	\end{bmatrix} + 
	\boldsymbol{D} \begin{bmatrix}
		\dot{\boldsymbol{x}}_{sp} \\
		\dot{\psi}_{sp}
	\end{bmatrix} + \boldsymbol{K}
	\begin{bmatrix}
		\boldsymbol{x}_{sp} - \boldsymbol{x}_{sp}^*\\
		\psi_{sp} - \psi_{sp}^*
	\end{bmatrix} =
	\begin{bmatrix}
		\hat{\boldsymbol{\delta}}_{{_f}} - \boldsymbol{\delta_{f}}_{des}  \\
		{\hat{\delta}}_{\tau}^{\boldsymbol{e_3}} - {\delta_{\tau_{des}}}
	\end{bmatrix}
\end{equation}
\textcolor{black}{where $\boldsymbol{x}_{sp}^*$ and $\psi_{sp}^*$ are reshaped into $\boldsymbol{x}_{sp}$ and $\psi_{sp}$ respectively by using the external wrench and yaw-torque information - $\boldsymbol{\hat{\delta}_f}$ and ${\hat{\delta}}_{{\tau}_{z}}$. The virtual parameters employed to reshape the trajectory are the virtual mass, $m_v \in \mathbb{R}^+$, inertia about $z$, $\mathcal{I}_{v,z}, \in \mathbb{R}^+$, and the virtual damping and spring gain matrices, $\boldsymbol{D} > 0 \in \mathbb{R}^{4\times4}$ and $\boldsymbol{K} > 0 \in \mathbb{R}^{4\times4}$.}

\subsubsection{Collision recovery}\label{sec:col_recovery}
The reference is generated by utilizing the current location and instantaneous collision force. Specifically, $\boldsymbol{x}_{sp} = \boldsymbol{x}_{t_c} + \mathcal{K}\boldsymbol{\hat{\delta}_f}$ with a user defined constant $\mathcal{K}$ and $t_c$ denotes the collision instant. The yaw setpoint is not modified. We employ this reaction mode to execute novel \textit{Ricocheting} maneuvers as will be described in Section \ref{sec:trj_Planning}. 

\section{Tactile-based Motion Primitives and Missions} \label{sec:contact mission}
\textcolor{black}{In this section, we present four primitives ($\Gamma \in \{1,2,3,4 \}$), for tactile-based exploratory and navigation missions by deformable UAVs. \revtwo{Unlike conventional motion primitives that rely on visual or inertial sensing, our tactile primitives enable navigation purely through contact interactions, making them highly effective in low-visibility or GPS-denied settings. We now state the contact dependent trigger conditions under which a primitive is autonomously engaged and the corresponding controllers from Section \ref{sec:int_ctrl} that become active.}}
\vspace{-0.1in}
\subsection{Case 1: Tactile-based Exploration and Mapping }\label{sec:tactile_exp}

\subsubsection{Tactile-based Exploration}\label{sec:exploration}
\rev{The exploration strategy is inspired by the coverage planning algorithms that combine random walk and wall tracing \cite{HR+14}. We introduce this scheme to aerial robots for the first time, demonstrating its effectiveness for deformable UAVs where compliance enables firm and stable contact, offering advantages over rigid UAVs.} 

\textcolor{black}{The tactile-exploration scheme involves flying forward until encountering an object, then maintaining contact to slide along its surface and turning around its edges to follow its contours. Towards this goal, we define two online decision variables: 
\begin{itemize}
	\item The contact direction, $\boldsymbol{C_n}\in \mathbb{R}^{4}$: A four-element vector specifying the contact-normal of an object in $\{\boldsymbol{b_1}, -\boldsymbol{b_1}, \boldsymbol{b_2}$ or $-\boldsymbol{b_2}\}$ directions, with $\|\boldsymbol{C_n}\|_1 = 1$. For instance, if $|{{{\hat\delta}}_f}^{\boldsymbol{\boldsymbol{b_1}}}| > {{{\delta}}}_{0}$ and ${{{\hat\delta}}_f}^{\boldsymbol{\boldsymbol{b_1}}} < 0$, $\boldsymbol{C_n} = [0~1~0~0]$, indicating an obstacle along $\{\boldsymbol{b_1}\}$ and the contact-normal along $\{-\boldsymbol{b_1}\}$. 
	\item The movement direction, $\lambda$: Selects movement direction from the set of $\{+X, +Y, -X$ or $-Y\}$, which corresponds to the $\{\boldsymbol{b_1}, \boldsymbol{b_2}, -\boldsymbol{b_1},  -\boldsymbol{b_2}\}$ directions as defined in Table I. 
\end{itemize}}
We now present the novel tactile-primitives for exploration (pseudo code in  \href{https://arxiv.org/pdf/2305.17217}{Suppl.} F, Algorithms 1-3).



\textbf{Exploration:} Initially, $\Gamma = 1$, corresponding to free-flight exploration \textcolor{black}{(referred hereon as simply \textit{Exploration})} during which XPLORER experiences negligible yaw torques and external forces. The controller from Section \ref{sec:free_flight} is utilized to generate the reference. We set ${d}_{step}$ to fly ``forward'' in the body frame, i.e $\lambda = +X$. The trigger condition for entering $\Gamma = 1$ at any time, is $(|\dot{\psi}| < \dot{\psi}_{0}$) and ($|{{\hat\delta^{\boldsymbol{\boldsymbol{b_1}}}_f}}|$ or ($|{{\hat\delta^{\boldsymbol{\boldsymbol{b_2}}}_f}}|)) < {{{\delta}}}_{0}$. 

\textbf{Tactile-turning:} \revtwo{The intuition is that upon encountering obstacles' external edges, the UAV should adjust its yaw to smoothly traverse outward corners allowing for continuous exploration.} Towards this, the yaw rate, $\dot{\psi}$, is used to detect a high turn rate around a point. This can occur when the vehicle slides along an object’s edge under a desired force, and then releases contact at a corner, causing a sudden spike in $\dot{\psi}$. Thus, if $|\dot{\psi}| > \dot{\psi}_0$, where $\dot{\psi}_0$ is the threshold, we set $\Gamma = 2$   
and activate the continuous yawing controller from Section \ref{sec:cont_yaw}, allowing the vehicle to rotate around a corner point while monitoring ${{{\hat\delta}}_f^{\boldsymbol{\boldsymbol{b_1}}}}$ (the force in $\{\boldsymbol{b_1}\}$ direction). If ${{{\hat\delta}}_f^{\boldsymbol{\boldsymbol{b_1}}}} > {{{\delta_{\psi_{0}}}}} $ at any time, indicating that further turning is not feasible, we transition to the next exploration state. 
The threshold $\dot{\psi}_0$ is chosen adequately to avoid false triggers and $\psi_{sp}$ is bounded to $[ -\pi ~ \pi]$ to avoid yawing indefinitely. 

\textbf{Tactile-traversal:} \revtwo{The intuition is that if the UAV encounters an object while flying forward, it should slide along its contour to adapt its path. Additionally, applying a controlled force while sliding will ensure stable contact for continuous wall-following and enable shape reconstruction. The sudden force release at edges or gaps will induce a high turn rate and autonomously trigger tactile-turning for rapid reorientation.} Accordingly, the forward contact force is monitored such that if $|{{\hat\delta^{\boldsymbol{\boldsymbol{b_1}}}_f}}| \text{ or } |{{\hat\delta^{\boldsymbol{\boldsymbol{b_2}}}_f}}| > {{{\delta}}}_{0}$, the vehicle is experiencing immovable force, setting $\Gamma = 3$. The \textit{Tactile-traversal} primitive utilizes controllers from Sections \ref{sec:free_flight} and \ref{sec:static_wrench} for controlled sliding. We set $\mathbf{\lambda}$ 
orthogonal to $\boldsymbol{C_n}$ following a right-side rule.

For instance, 
if $\boldsymbol{C_n}=[0~1~0~0]$,  
$\lambda = +Y$ and the next waypoint is first generated utilizing the controller in Section \ref{sec:free_flight} with ${d}_{step}$ for motion in $+Y$. The interaction controller from Section \ref{sec:static_wrench} then reshapes it to ensure that the vehicle maintains contact with the obstacle by exerting a desired force on it while sliding across it, as shown in Fig. \ref{fig:first_pic}(ii). Specifically, $\boldsymbol{\delta_{f}}_{des} = [\Delta_{f_{des}} ~0~ 0]^T$ and $\delta_{\tau_{des}} = \hat{\delta}_{\tau}^{\boldsymbol{e_3}}$ to enable force exertion in $+X$, flying in $+Y$ and yielding to any yaw torques.

\textit{Remark 1:} ${{{\delta_{\psi_{0}}}}}$ is chosen slightly higher than ${{{\delta}}}_{0}$ to ensure fail-safe behavior while transitioning from $\Gamma=2$ to $\Gamma =3$.

\subsubsection{Tactile-based Mapping}
\label{sec:mapping}
\begin{figure}
\vspace{-0.1in}
\centering
\subfloat{\includegraphics[trim = 0.1 0.1cm 0 0.1cm, clip, width = 0.24\textwidth]{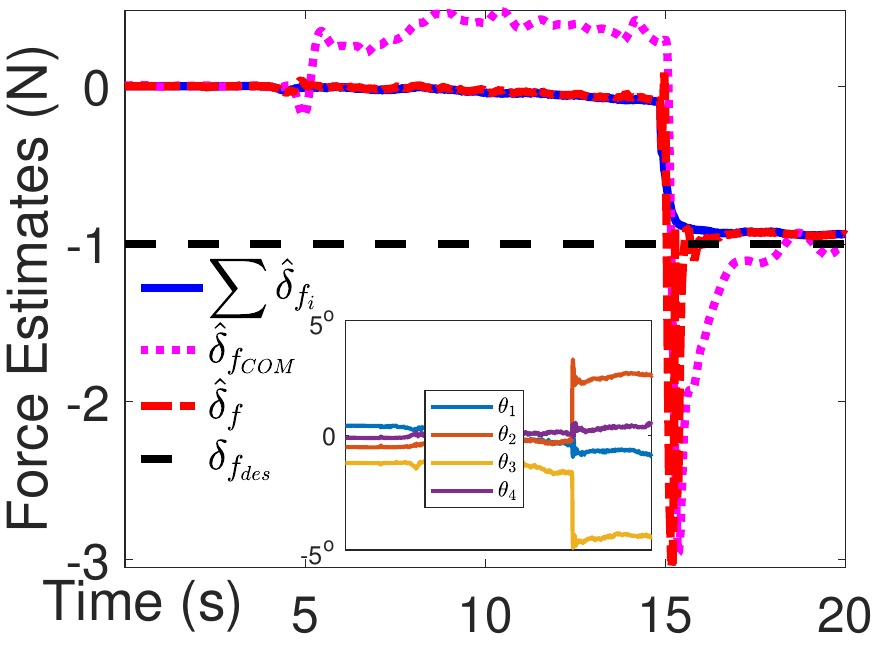}}
\subfloat{ \includegraphics[width = 0.24\textwidth, height = 0.179\textwidth, trim = 0 0.1cm 0 0.1cm, clip]{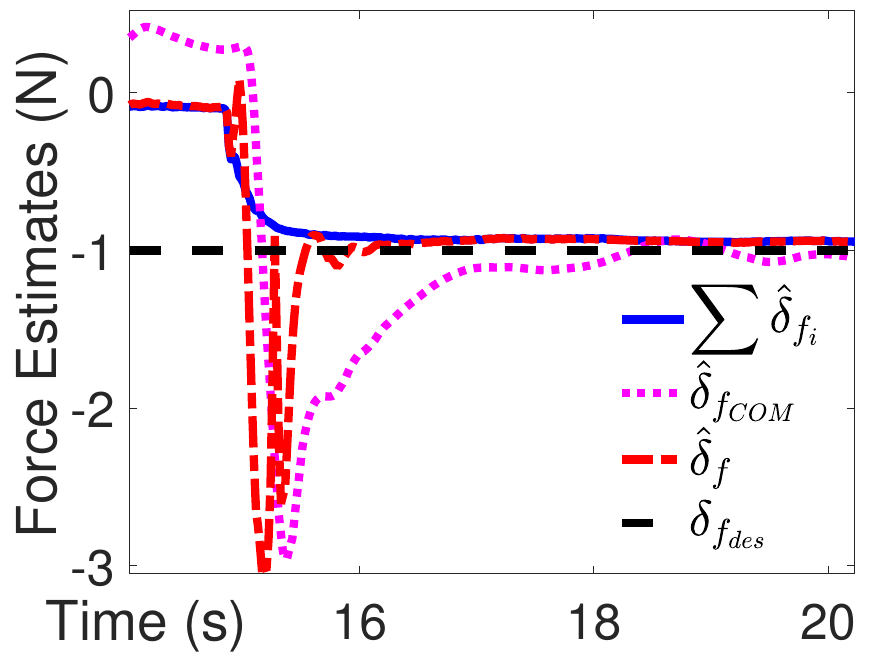}}
\vspace{-0.1in}
\caption{Results for the proposed force estimation algorithm. Left: Shows estimates for $\boldsymbol{\hat{\delta}_f},\boldsymbol{\hat{\delta}_{CoM}}$ and $\sum \boldsymbol{\hat{\delta}_{f_i}}$; inset shows the corresponding deviation in arm angles. Right: zoomed-in view of the peak. 
}
\vspace{-0.2in}
\label{fig:wrench_results2}
\end{figure}
The ability to explore the environment while maintaining contact can be utilized to generate an obstacle map. This map can then be used for motion planning by XPLORER or other autonomous robots. The algorithm utilizes the $ {\boldsymbol{\hat\delta_f}} $, the CAD model of XPLORER, and the global pose to generate the obstacle boundary map. We use the Open3D library\cite{ZP18} to process the point cloud generated. 

\rev{Obstacle map generation initializes if the estimated force is above the threshold, i.e., } \rev{$ |{{\hat\delta^{\boldsymbol{\boldsymbol{b_1}}}_f}}| $ or $ |{{\hat\delta^{\boldsymbol{\boldsymbol{b_2}}}_f}}|  \ge {\delta_{map}}$} and if the UAV is flying. ${\delta_{map}}$ is taken to be slightly higher than $f_{des}$ to ensure mapping is conducted only when there is firm contact with the obstacle. A rigid object of dimension 0.25 m $\times$ 0.08 m $\times$ 0.5 m is added to the point cloud at a small offset from the CoM of XPLORER in the direction of $\boldsymbol{C_n}$, which provides the outward normal of the obstacle for shape reconstruction. 

To handle internal corners, we utilize the previous and current move directions to determine object placement. Specifically, when the move direction, $\lambda$ differs from the previous one, it typically indicates a corner, prompting the addition of a fixed point cloud diagonally with a 0.417 m offset to ensure map continuity. These parameters were selected to suit indoor environments and improve mapping accuracy.

The point cloud map is generated at a frequency of 30 Hz to achieve a high-resolution representation of the obstacle. It is stored in Polygon File Format (PLY) in ASCII format, containing the coordinates of each point in the cloud. The map generation process was carried out on a computer equipped with an AMD Ryzen 5 CPU and 16GB of RAM. The position and wrench estimates are recorded on the ground computer and relayed from the high-level companion computer via ROS2 network. An overview of the framework is in \href{https://arxiv.org/pdf/2305.17217}{Suppl.} Alg. 4. 

\vspace{-0.1in}
\subsection{Case 2: Tactile-based Navigation - Ricocheting}\label{sec:trj_Planning}
We now introduce \textit{Ricocheting} -- a tactile-based primitive for fast braking by bouncing-off objects. 

\begin{figure}
\centering
\subfloat{\includegraphics[ width = 0.24\textwidth]{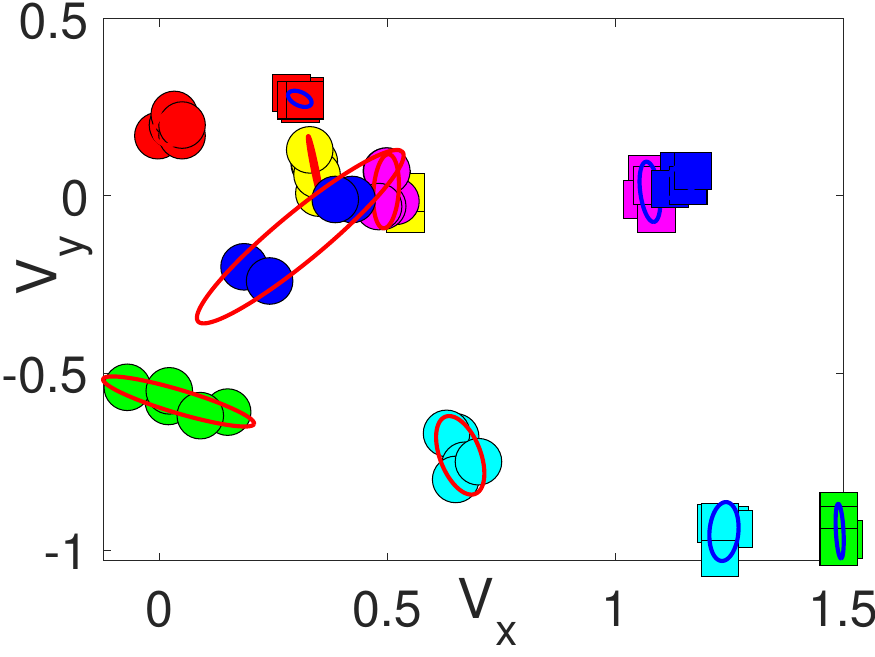}} 
\subfloat{\includegraphics[trim = 0 0 0 0.7cm, clip, width=0.24\textwidth]{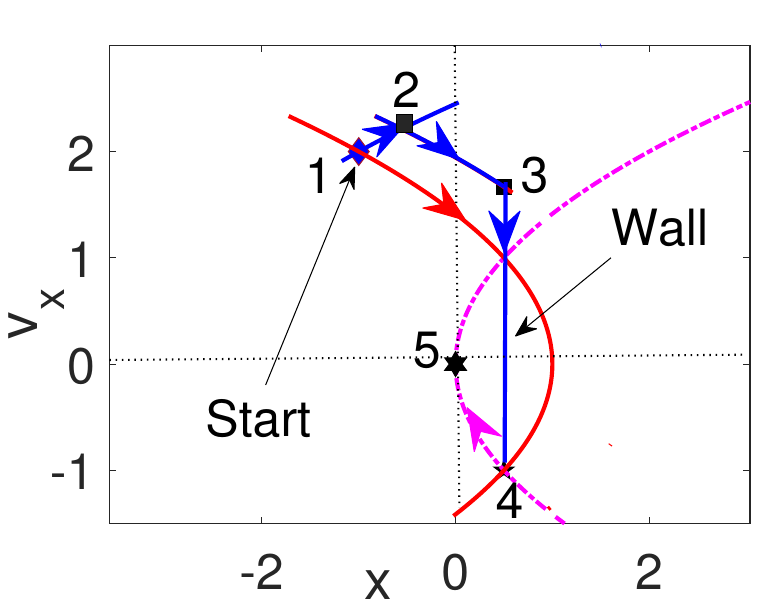}}
\vspace{-0.1in}
\caption{Left: Collision experiments to understand ricocheting. Symbols $\square$  and the $\circ$ represent the pre and post-collision velocities of each case (distinguished by color). 
	Right: Simulation phase-portrait to demonstrate ricocheting (path 1-2-3-4-5) vs conventional (path 1-4-5). The curves represent constant acceleration for a second order system, $\Ddot{x} = u$, with $u$ as the input.}
\label{fig:collision_NN}
\vspace{-0.2in}
\end{figure}

\textcolor{black}{First, we collected experimental data for six different collision velocities with two different heading angles (at $\psi$ = 90$\degree$, 45$\degree$) with five trials each, totaling 60 flights to understand the post-collision dynamics. In all the experiments, the recovery controller from Section \ref{sec:col_recovery} is employed to generate the post collision waypoint. The scatter plots for these experiments are shown in Fig. \ref{fig:collision_NN}, along with the error ellipses. We noticed that the heading angle didn't influence the post collision state significantly. Moreover, in all the experiments, there was energy loss due to the damping introduced by the deformable chassis resulting in dissipation of energy which combined with the recovery controller, facilitates a fast stopping maneuver.} 

This enables us to define \textit{Ricocheting} ($\Gamma =4$), as a maneuver where collisions between the UAV and environment serve as jump maps to facilitate minimum-time trajectories. Following the time-optimal control approach for hybrid double integrators with state-driven jumps in \cite{AC+19}, simulation results in Fig. \ref{fig:collision_NN}(b) show that the path 1-2-3-4-5 is shorter than 1-4-5 when state jumps are allowed (the complete derivation is in \href{https://arxiv.org/pdf/2305.17217}{Suppl.} G). However, the UAV ricocheting problem differs from the setting in \cite{AC+19} because ``collisions'' serve as jump maps, which cannot be activated arbitrarily at any location. Therefore, we developed a neural network (CollisionNet) to analyze the rebound velocities for the data collected (details in \href{https://arxiv.org/pdf/2305.17217}{Suppl.} H) and propose the condition whether or not to ricochet. 

\textit{When to Ricochet}: If a collision node lies within a circle of radius $\zeta$ of the stopping node (marked in Fig. \ref{fig:first_pic}(a3)), ricochet off that node at maximum velocity attainable to reach the stopping node in minimum time. The post-collision velocities can be verified online via the CollisionNet developed before engaging $\Gamma = 4$. We choose $\zeta = 0.5$m.


\revtwo{\textit{Remark 2}: In most experiments, collisions introduced an undesired downward force, so we avoid exploiting momentum changes along $\boldsymbol{e_3}$ and focus only on in-plane (2D) collisions (hence a circle in the above condition). Moreover, as heading angle changes have a minor impact on momentum relative to the pre-collision velocity, any 2D maneuver can be effectively reduced to 1D for multirotors, ensuring the condition is valid.}

\revtwo{\textit{Remark 3:} Significant variation in the post collision state at higher velocities (large blue and green scatter groups in Fig. \ref{fig:collision_NN}) warrant thorough collision modeling to better understand impact dynamics. Such modeling is essential for proposing new maneuvers at higher velocities, beyond those introduced in this article.}

\section{Experimental Results and Discussion}\label{sec:experiments}
\subsection{Experimental Setup, Sensors and Control Parameters}\label{appx:electronics}
\subsubsection{Experimental Setup} The low-level controller utilized is a PIXHAWK flight controller with the RaspberryPi4B as the companion computer. The companion computer relays the position and orientation of the vehicle from an indoor motion capture system to the flight controller at 120 Hz. A 4S lithium polymer battery of 3300 mAh LiPo battery of 14.8V, 50C is used for the power supply. The motors are controlled utilizing Lumenier 30A BLHeli\_S Electronic Speed Controllers and the entire system has a mass of 1.12 kg. 
Our experimental testbed consists of four distinct environments that closely resemble real-world flight spaces and help in validating the tactile-based exploration and navgiation algorithms. (i) a rectangular object measuring 1.22 m $\times$ 1.0 m was set up using acrylic panels (ii) acrylic panels were organized to represent a box with XPLORER's initial state outside the box (iii) a trash can and (iv) a vertical pipe to represent real obstacles.

\subsubsection{Sensors for Arm-based Force Estimates}

\rev{We employ four low-cost off-the-shelf 9-DOF IMUs (BNO055, Adafruit, New York, NY) on XPLORER (one on each arm) to get the angular acceleration and estimate the external force on the arm. They are connected to the companion computer via serial communication at 50 Hz.} The Euler angles are computed using the Adafruit BNO055 library. A median low-pass with a band-stop filter is employed to obtain accurate estimates of the arm angles, $\theta_i,$ at any given instant. This is used to estimate the arm wrench $\boldsymbol{\hat{\delta}_{f_i}}$ in Section \ref{sec:arm_wrench}.  

\begin{table}
	\centering
	\caption{Parameters \& Thresholds used in Experiments}\label{table:threshold}
	\vspace{-0.1in}
	\begin{tabular}{|c|c||c|c|}
		\hline 
		\textbf{Parameter} & \textbf{Threshold Value} & \textbf{Parameter} & \textbf{Threshold Value} \\
		\hline
		$ \dot{\psi_0} $  & 0.4 rad/s & $ d_{step}$ & 0.25 m\\
		$ \mathcal{W} $  & 0.26 rad/s & $ \Delta_{f_{des}}$ & 1.25 N  \\
		$ {\delta}_0 $ & 1.5 N & $ {\delta}_{map} $ & 1.51 N\\
		${{\delta_{\psi}}}_{0}$ & 1.6 N & $\mathcal{K}$ & 0.1\\
		\hline
	\end{tabular} 
	\vspace{-0.2in}
\end{table}
\subsubsection{Thrust Estimate and Interaction Controller Parameters}\label{sec:wrench_exp}
\rev{In order to calculate the wrench at the CoM, the controller's normalized force and torque values are obtained from the onboard flight controller. The force value is obtained empirically and used to calculate the approximate value of the actual thrust. This technique can be improved by performing a PWM-thrust mapping curve using RPM sensor feedback. Gains for the interaction controller in (\ref{eqn:interaction_controller}) are tuned by experiments to be: $\boldsymbol{D} = \boldsymbol{K} = diag(24.5,24.5,0,1), m_v = \mathcal{I}_{v,z} = 1 $}.
\vspace{-0.1in}
\subsection{Force Estimation Algorithm Validation} A static wrench application task is performed to validate the force estimation algorithm as shown in Fig. \ref{fig:wrench_results2}. The reference for this experiment is set to $\boldsymbol{\delta_f}_{{des}} = [-1~0~0]^{T}$ for applying a 1 N force on the wall. We use a load cell (ATO Micro 5kg Tension and Compression Load Cell, S Type, ATO, Diamond Bar, CA) with a sampling rate of 20Hz, mounted on the wall, to obtain the ground truth. A digital reader is attached to the load cell which is calibrated to display the force readings when the load cell is compressed. The UAV takes off and starts applying force on the box whose other end pushes the load cell. Wrench validation experiments are shown in Supplementary Video, SVideo (Part2). 

The inset plot in Fig. \ref{fig:wrench_results2}, shows the deviation of the arm angles upon contact. The estimate from the CoM (magenta dotted line) takes time to converge, however, the arm-based force estimate (blue solid) converges sooner without the collision peak. By fusing both methods, the proposed estimate (red dashed) retains the peak from the CoM-based method and also converges faster. \textcolor{black}{For this experiment, the true force applied on the wall was collected and averaged over 2s after impact, to be approximately 1.31N, demonstrating an accuracy of $\approx 77\%$.} We also conduct experiments where there is no contact to show $\boldsymbol{\hat{\delta}_f} = \boldsymbol{\hat{\delta}_f}_{CoM}$ (SVideo Part2, Suppl. E, pulley experiment). 

\vspace{-0.1in}
\subsection{Autonomous Tactile-based Exploration and Mapping}
\begin{figure}[t]
	\centering
	\subfloat[\revtwo{Wall traversal by XPLORER, successful in all three trials} \label{fig:Wall_Traversal_Graph}]{\includegraphics[width = 0.45\textwidth]{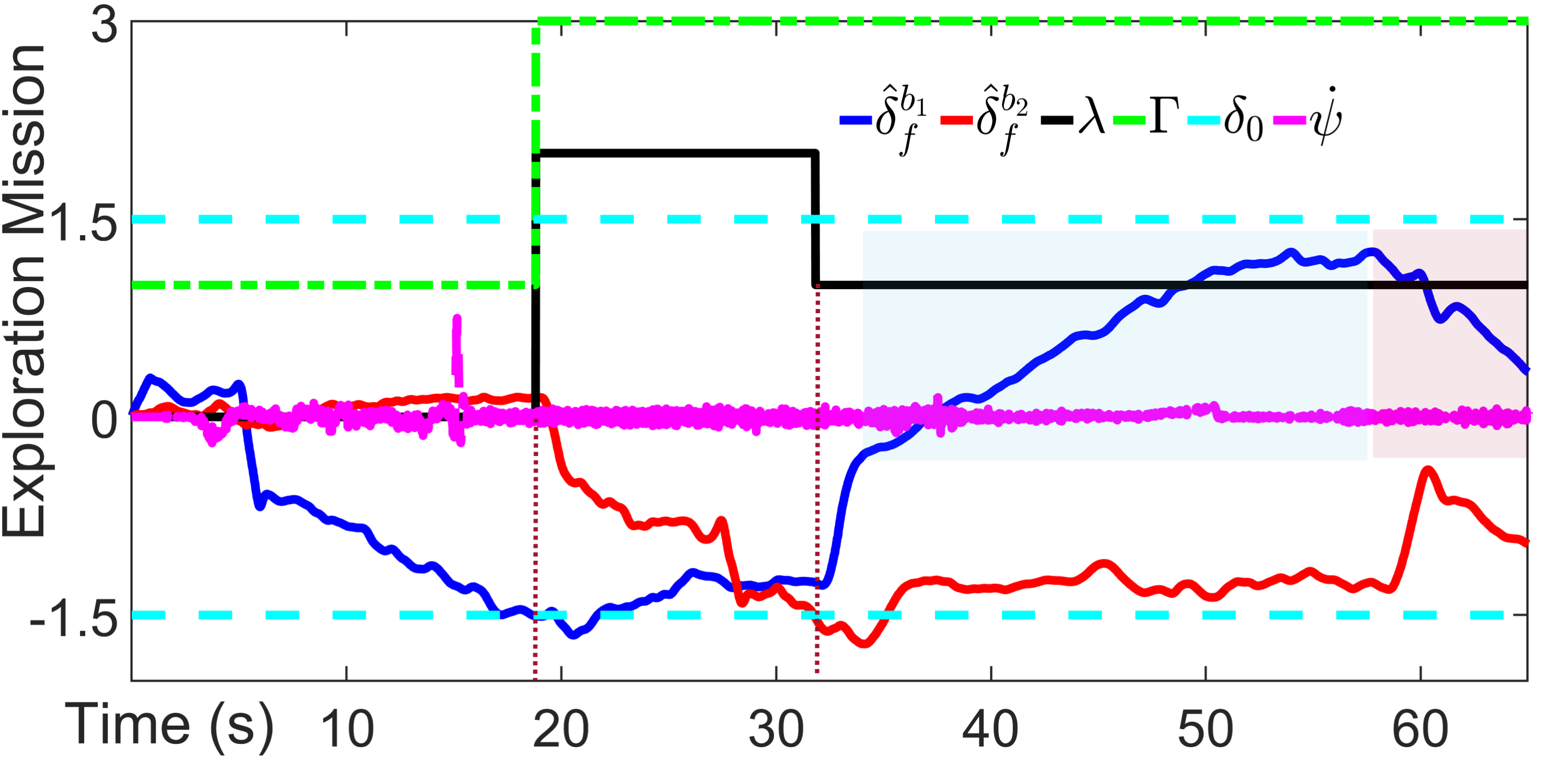}}\\
	\vspace{-0.1in}
	\subfloat[\revtwo{Box traversal by XPLORER, successful in all three trials.}\label{fig:Box_Traversal_Graph}]{\includegraphics[width = 0.45\textwidth]{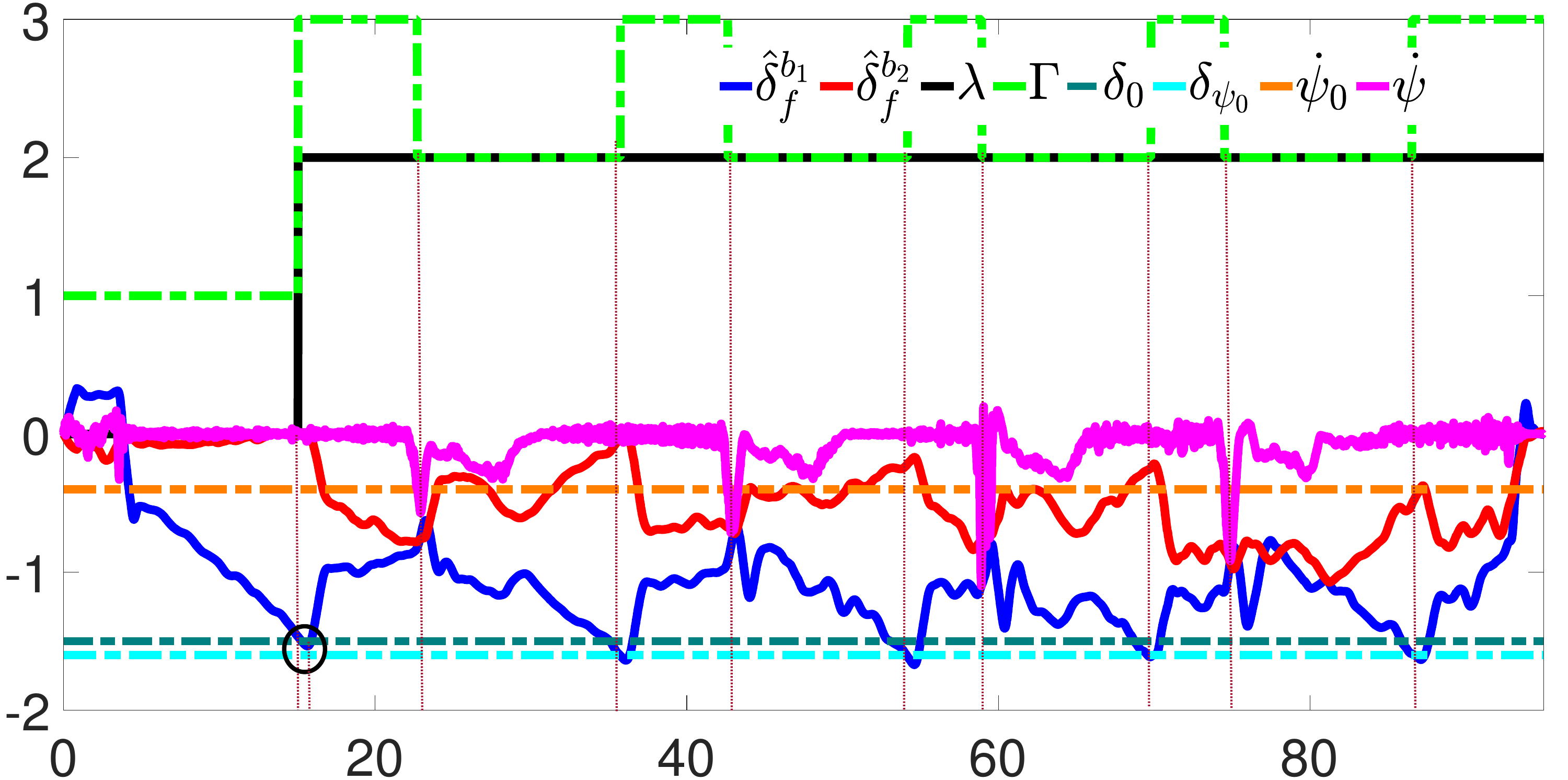}}\\
	\vspace{-0.1in}
	\subfloat[\revtwo{Box traversal by a rigid UAV, failed in all three trials} \label{fig:Traversal_Graph_Rigid}]{\includegraphics[width = 0.45\textwidth]{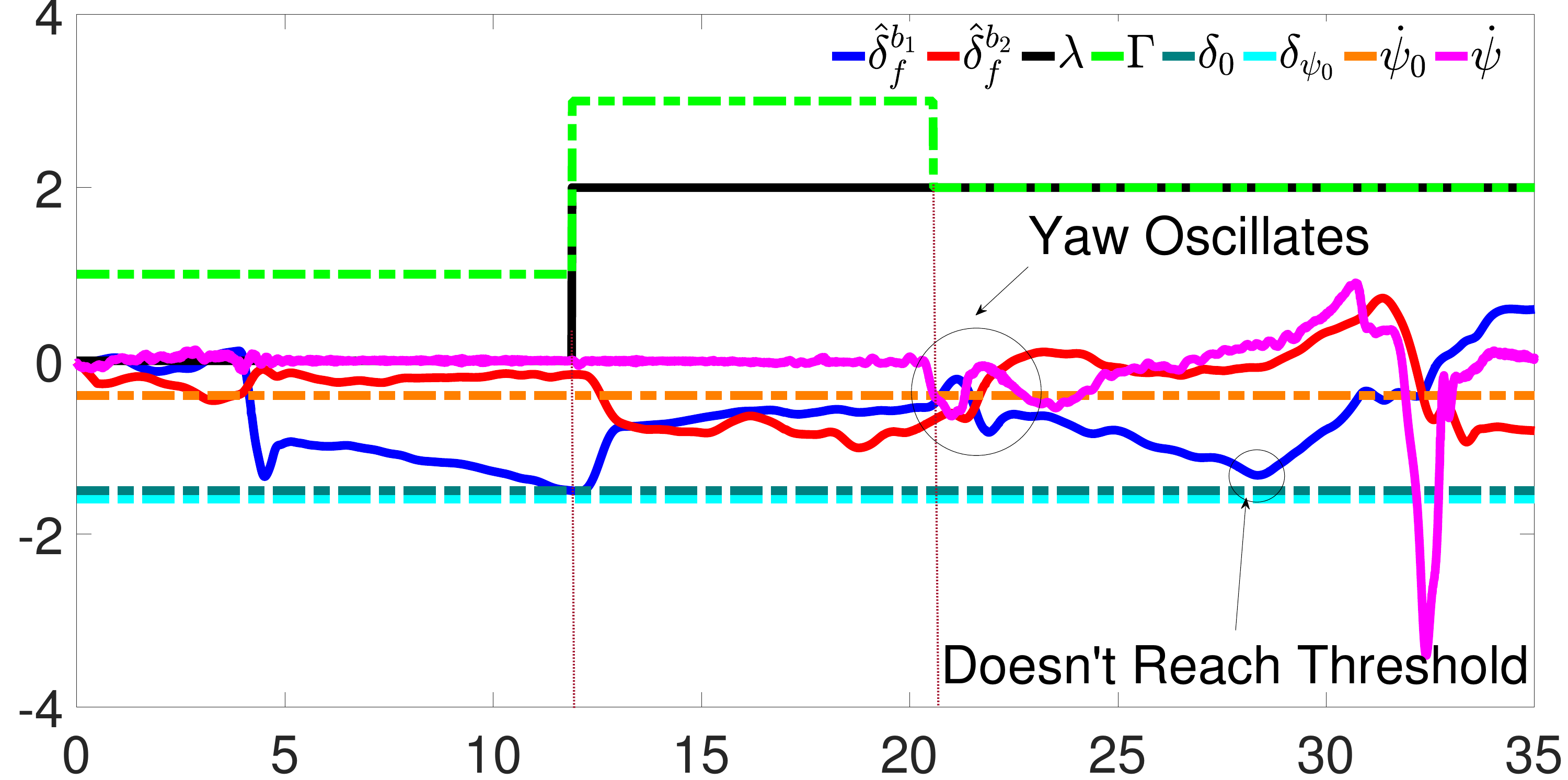}}
	\caption{Experimental results for tactile-based exploration. (i) XPLORER navigating a wall-like structure, activating $\Gamma=$1,3 when $\delta_0$ is crossed. Blue and red regions indicate unmodeled frictional and wall effects. (ii) XPLORER navigating a box-like structure, activating $\Gamma=$1,2,3 when $\delta_0$ or $\dot{\psi}_0$ are crossed. (iii) A rigid UAV fails to engage Tactile-traversal after Tactile-turning due to chassis rigidity, resulting in oscillatory yaw motion at convex corners.}
	\label{fig:exploration_results}
	\vspace{-0.2in}
\end{figure}

\begin{figure}[t] 
	\centering
	\subfloat{\includegraphics[trim = 8.5cm 1cm 0 0, clip, width = 0.18\textwidth]{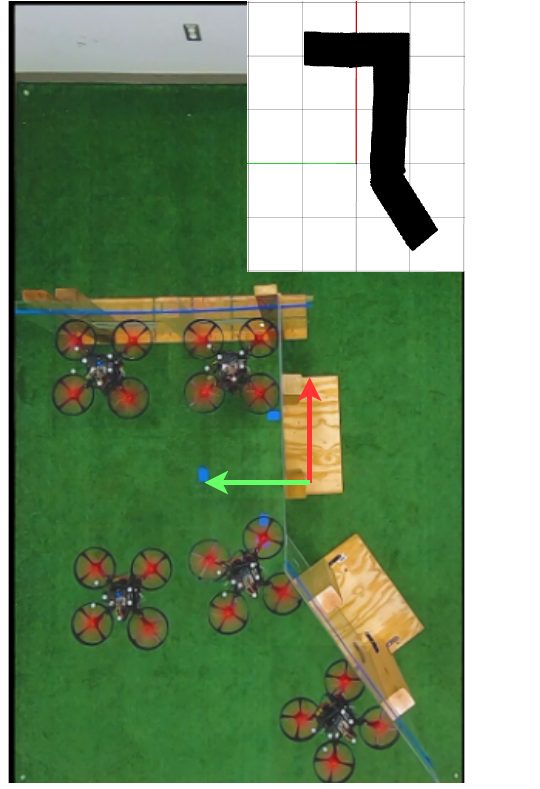}} \hspace{-0.2in}
	\subfloat{\includegraphics[trim = 0 1cm 9cm 0, clip, width = 0.17\textwidth]{figures/Top_View_R4.pdf}}\hspace{-0.13in}
	\subfloat{\includegraphics[width = 0.16\textwidth, trim = 0 1cm 10cm 0, clip]{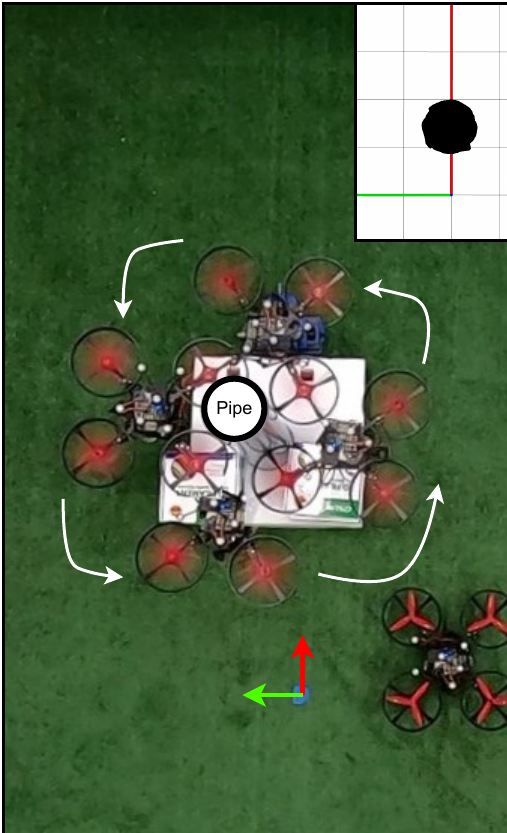}}
	\vspace{-0.1in}
	\caption{Top view of the mapping experiments with the corresponding maps generated, in the insets for a box, a wall and an acrylic pipe.}
	\label{fig:Aerial_View}
	\vspace{-0.3in}
\end{figure}

In this section, we present the experimental results for the exploration and mapping mission applied to map a wall, a box, an acrylic pipe, and a trashcan as shown in Figs. \ref{fig:exploration_results} and \ref{fig:Aerial_View}. 

The mission begins with XPLORER taking off to a hover height of 0.7m and initiating the \textit{Exploration} state ($\Gamma = 1$), moving forward until an obstacle is detected. It then pushes against the surface until the threshold for \textit{Tactile-traversal} is reached, triggering a transition to \textit{Tactile-traversal} ($\Gamma = 3$), where it slides along the obstacle’s surface. This $1 \rightarrow 3$ transition is illustrated in Fig.~\ref{fig:Wall_Traversal_Graph}.

If the vehicle encounters an outward corner (e.g., on a box) that induces a sudden yaw, it transitions to \textit{Tactile-turning} ($\Gamma = 2$) to perform a controlled turn and establish contact with the next surface, before returning to \textit{Tactile-traversal} ($\Gamma = 3$). For box-like structures, the full state sequence becomes $1 \rightarrow 3 \rightarrow 2 \rightarrow 3$, as shown in
\ref{fig:Aerial_View}.

\revtwo{\textit{Remark 4:} Performance of the exploratory algorithm relies on threshold values which are influenced by friction and aerodynamic wrenches acting on XPLORER. Through extensive testing, we identified threshold values that perform reliably in indoor environments, as shown in Table \ref{table:threshold}. While the compliant chassis allows these parameters to generalize across a broad range, accurately updating the desired interaction wrench remains challenging and is an active area of research.}

\textit{Note:} The state machine applies a moving average filter (50 samples) on ${\boldsymbol{\hat{\delta}_f}}$, and a low-pass filter on $\dot{\psi}$ to mitigate noise effects during decision-making.

\subsubsection{Wall-traversal}
In this experiment, XPLORER navigates across a wall edge. $\Gamma$ switches between 1 and 3 only as shown by the green line in Fig. \ref{fig:Wall_Traversal_Graph} and SVideo (Part3a); the yaw-rate never crosses $\dot{\psi}_0$ for \textit{Tactile-turning} to be engaged. The mapping algorithm follows the description in Section \ref{sec:mapping} and the generated map is shown in the inset of Fig. \ref{fig:Aerial_View}. 
\subsubsection{Box-traversal} In this experiment, XPLORER navigates around a box. From Fig.~\ref{fig:Box_Traversal_Graph}, $\Gamma$ correctly switches between values of 1, 2, and 3 whilst circumnavigating the obstacle. The box dimensions extracted from the generated map (inset of Fig.~\ref{fig:Aerial_View}) are 1.231m$\times$1.019m, compared to the actual dimensions of 1.22m$\times$1.0m, yielding an area estimation accuracy of  $\approx$96.72\%. The results are demonstrated in SVideo (Part 3b).
\subsubsection{Pipe outer diameter traversal} This test case demonstrates the tactile-based exploration algorithm's capability to navigate circular objects. The generated map in the inset of Fig. \ref{fig:Aerial_View} and SVideo (Part3c) for this particular scenario, is within 86\% of the actual dimensions of the pipe. Compared to the previous scenario, the drop in mapping accuracy is possibly due to instances when XPLORER momentarily loses contact with the pipe before re-establishing contact. 
\subsubsection{Trash can traversal} This case presents a partial success scenario when during tactile-based exploration, XPLORER loses contact (final edge of the trash can in SVideo (Part3c)). \revtwo{The combination of a low-friction surface, the can’s size relative to XPLORER’s width, and insufficiently tuned virtual parameters} in (\ref{eqn:interaction_controller}) likely results in inadequate $\dot{\psi}$ at the corner, preventing the system from transitioning into the \textit{Tactile-turning} state. The mission is manually terminated here. In the future, design modifications and online tuning of controller parameters can be looked into to prevent such failures.




\vspace{-0.15in}
\rev{\subsubsection{Comparison against a Rigid UAV}\label{appx:rigid_vs_explorer_exploration}
	We conducted both wall-traversal and box-traversal experiments for a similar dimension rigid UAV and only the wall-traversal exploration with inward edges was successful. For a box-like obstacle, the rigid UAV initially is successful in engaging $\Gamma$ as $1 \rightarrow 3 \rightarrow 2$ systematically as shown by the green line in Fig. \ref{fig:Traversal_Graph_Rigid}. However, after the turning maneuver in $\Gamma = 2$, significant oscillations are observed in yaw and the UAV was not able to maintain consistent contact, possibly due to absence of compliance and damping in the chassis. Consequently, $\Gamma = 3$ was never engaged as compared to the XPLORER case, where reengagement happened around 37s as shown in Fig.~\ref{fig:Box_Traversal_Graph}. Similar behaviors are observed in all four experiments conducted for this case, and the exploration scheme for a rigid drone failed. Experimental videos are presented in SVideo (Part3d).}

\textit{Note:}  Additional validation results are presented in SVideo Part3c for objects with slots. It is also observed that in some cases when the rigid UAV makes contact in \textit{Exploration} state, significant yaw torque disturbance (due to the large impacts) falsely triggered the \textit{Tactile-turning} state, reinforcing that the proposed tactile-based exploration scheme is suitable for deformable UAVs. These results are presented in \href{https://arxiv.org/pdf/2305.17217}{Suppl.} I.

\begin{figure}[t]
\centering
\subfloat{\includegraphics[width = 0.24\textwidth]{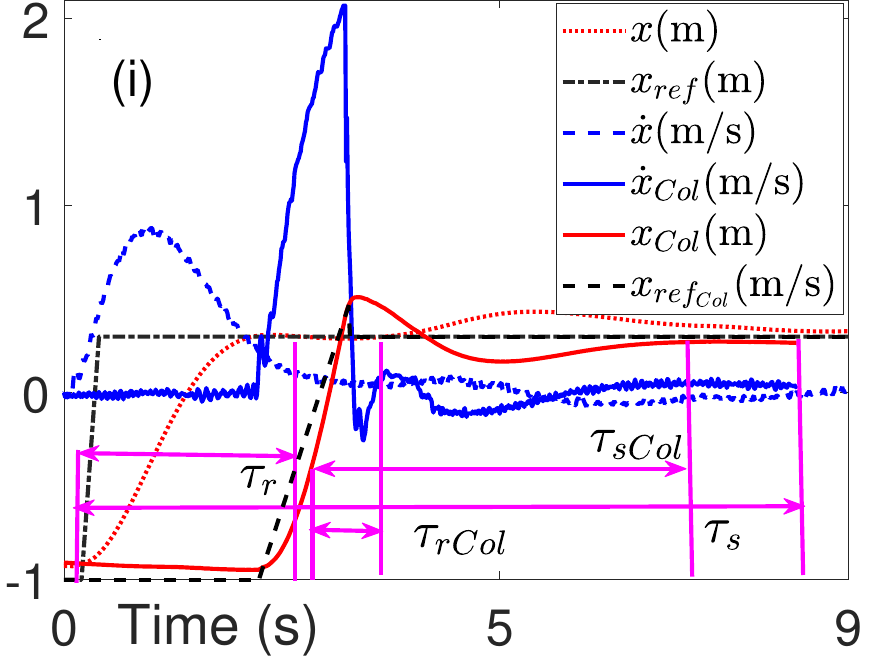}}
\subfloat{\includegraphics[width = 0.24\textwidth]{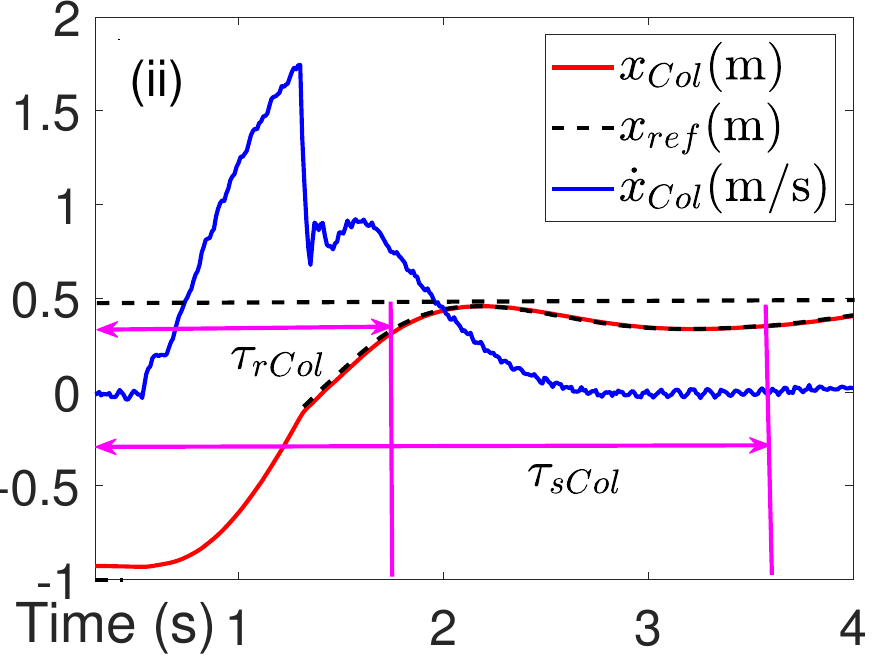}}
\vspace{-0.1in}
\caption{Trajectories of three ricocheting experiments. (i) Goal is near the wall, shows both with- and without- ricocheting. (ii) Goal location is beyond wall, ricocheting off the edge. In all cases, ricocheting maneuvers are time-optimal.}
\label{fig:results_exp_mp}
\vspace{-0.2in}
\end{figure}
\vspace{-0.1in}
\subsection{Waypoint Tracking via Ricocheting}\label{sec:ricocheting_results}
In this subsection, we present the benefits of ricocheting via two cases. In the first case, the goal ($[0.3 ~0~-0.75]^T$m) is near the wall ($x_{wall} = 0.5$m) to facilitate \textit{ricocheting off the wall's face} with a velocity of $[\dot{x}^-,\dot{y}^-]^T = [2,~0]^T$m/s. In the second case, the goal ($[0.5 ~0~-0.75]^T$m) is set some distance beyond the wall ($x_{wall} = 0$m) to enable \textit{ricocheting off the edge} with $[\dot{x}^-,\dot{y}^-]^T = [1.75,~0]^T$m/s. For both the cases, the start location is $[-1 ~0~-0.75]^T$m. The pre-collision velocities were the maximum velocities that could be attained in the closed space and are verified by the CollisionNet in Section. \ref{sec:trj_Planning} to have reduced rebound velocities upon impact. 
The rise time ($\tau_r$) and settling time ($\tau_s$) are chosen as the metrics to evaluate the performance of ricocheting. 

For the first case scenario with the goal near the wall, the conventional, collision-exclusive trajectory is shown in Fig. \ref{fig:results_exp_mp}i, where the vehicle first accelerates, then decelerates to reach the setpoint. There is an overshoot which it tries to minimize, slowly converging to the reference. The $\tau_r$ and $\tau_s$ are noted to be 2.27 and 8.995 seconds, respectively. Furthermore, over three trials the RMSE error was around $\pm$ 3cm for this case. In contrast, the experimental results for ricocheting show that XPLORER reaches the wall with maximum velocity, collides, and stops at the wall almost instantaneously by dissipating kinetic energies as shown by the plots in Fig. \ref{fig:results_exp_mp}i. The $\tau_{rCol}$ (subscript $_{Col}$ denotes experiments with collision) and $\tau_{sCol}$ are calculated to be 1.046 and 5.292 seconds, respectively which are faster than the conventional trajectories. Over six trials, the RMSE error was $\pm$ 0.5cm. 

For the second case where the vehicle performs ricocheting off the edge and regulates itself to a setpoint beyond the wall, results in Fig. \ref{fig:results_exp_mp}ii show significantly shorter flight time with $\tau_{rCol} = 1.685$ seconds and $\tau_{sCol} = 3.631$ seconds, respectively.  The flight test results are shown in SVideo (Part4).

For both cases, it is seen that \textit{Ricocheting} with the current recovery controller yielded agile maneuvers with shorter flight time by incorporating a collision-based velocity state-jump. This highlights the advantages of collisions for planning novel pursuit/evade maneuvers and agile navigation.

\section{Conclusion}\label{sec:conclusion}
\subsection{Summary}
\revtwo{In this work, we presented an autonomous exploration and navigation framework for deformable UAVs through real-time trajectory modifications. Unlike conventional motion primitives, our tactile primitives adapt dynamically to surface interactions, improving navigation and tactile mapping. Experiments showed that tactile-turning allows UAVs to execute smooth corner traversal, while tactile-traversal ensures continuous surface contact for reliable wall-following. Additionally, our force estimation algorithm effectively triggered these primitives, enabling robust autonomous decision-making. We also introduced ricocheting, leveraging collision-induced state jumps for minimum-time braking and trajectory redirection.} 

\revtwo{The proposed tactile-based exploration and navigation methodology demonstrates that deformable UAVs can interact with their environment in ways rigid UAVs cannot, unlocking new capabilities for autonomous navigation, inspection, and surveillance in constrained, contact-rich environments.}

\vspace{-0.1in}
\subsection{Discussion and Future Directions}
\revtwo{Future research can focus on developing detailed collision models to enable new maneuvers in 3D space beyond those proposed in this article. These behaviors could be integrated into online motion planning algorithms, such as RRT*, for improved real-time trajectory adjustments. Another direction is to use learning-based strategies to adapt the virtual stiffness ($\boldsymbol{K}$) and damping ($\boldsymbol{D}$) parameters and exploration step size ($d_{step}$) based on surface properties for robust interaction. Incorporating tactile sensing with visual-inertial odometry could strengthen navigation in GPS-denied or low-visibility settings. Finally, advanced Kalman filtering techniques could be explored to fuse proprioceptive data for wrench estimates, improving accuracy for various conditions. Together, these directions could significantly broaden the capabilities of UAVs for physical interaction in complex environments.}

\section{Acknowledgments}
The authors thank YiZhaung Garrard, Bill Nguyen and Yogesh Kumar from the ASU RISE lab for the brainstorming sessions and help with experiments.
\bibliographystyle{ieeetr}
\bibliography{bibliography.bib}

\begin{thebibliography}{10}

\bibitem{RL18}
F.~Ruggiero {\em et~al.}, ``Aerial manipulation: A literature review,'' {\em
  IEEE Robot. Autom. Lett}, vol.~3, no.~3, pp.~1957--1964, 2018.

\bibitem{HZ+05}
D.~Hausamann {\em et~al.}, ``Monitoring of gas pipelines--a civil uav
  application,'' {\em Aircraft Eng. Aero. Tech.}, vol.~77, no.~5, pp.~352--360,
  2005.

\bibitem{MS+21}
S.~Mishra {\em et~al.}, ``Autonomous vision-guided object collection from water
  surfaces with a customized multirotor,'' {\em IEEE/ASME Trans. Mech.},
  vol.~26, no.~4, pp.~1914--1922, 2021.

\bibitem{KC+13}
S.~Kim {\em et~al.}, ``Aerial manipulation using a quadrotor with a 2dof
  robotic arm,'' in {\em IEEE/RSJ Int. Conf. Intl. Robot. Sys.},
  pp.~4990--4995, 2013.

\bibitem{SH+15}
A.~Suarez {\em et~al.}, ``Lightweight compliant arm for aerial manipulation,''
  in {\em IEEE/RSJ Int. Conf. Intl. Robot. Sys.}, pp.~1627--1632, 2015.

\bibitem{B+24}
A.~Bredenbeck {\em et~al.}, ``Embodying compliant touch on drones for aerial
  tactile navigation,'' {\em IEEE Robot. Autom. Lets}, vol.~10, no.~2,
  pp.~1209--1216, 2024.

\bibitem{K16}
M.~Kovac, ``Learning from nature how to land aerial robots,'' {\em Science},
  vol.~352, no.~6288, pp.~895--896, 2016.

\bibitem{MF16}
S.~Mintchev and D.~Floreano, ``Adaptive morphology,'' {\em IEEE Robot. Autom.
  Mag}, vol.~23, no.~3, pp.~42--54, 2016.

\bibitem{TP23}
W.~Tao {\em et~al.}, ``Design, characterization and control of a whole-body
  grasping and perching ({WHOPPE}r) drone,'' in {\em IEEE/RSJ Int. Conf. Intl.
  Robot. Sys.}, pp.~1--7, IEEE, 2023.

\bibitem{PW23}
K.~Patnaik and W.~Zhang, ``Adaptive attitude control for foldable quadrotors,''
  {\em IEEE Control Systems Letters}, 2023.

\bibitem{PW21}
K.~Patnaik and W.~Zhang, ``Towards reconfigurable and flexible multirotors,''
  {\em Int. J. Intel. Robot. App.}, vol.~5, no.~3, pp.~365--380, 2021.

\bibitem{PM+20}
K.~Patnaik {\em et~al.}, ``Design and control of squeeze,'' in {\em IEEE/RSJ
  Int. Conf. Intl. Robot. Sys.}, pp.~1364--1370, 2020.

\bibitem{LK21}
Z.~Liu and K.~Karydis, ``Toward impact-resilient quadrotor design, collision
  characterization and recovery control to sustain flight after collisions,''
  in {\em IEEE Int. Conf. Robot. Autom.}, pp.~183--189, 2021.

\bibitem{PM+21}
K.~Patnaik {\em et~al.}, ``Collision recovery control of a foldable
  quadrotor,'' in {\em IEEE/ASME Int. Conf. Adv. Intl. Mech.}, pp.~418--423,
  2021.

\bibitem{PK+23}
P.~H. Nguyen, K.~Patnaik, {\em et~al.}, ``A soft-bodied aerial robot for
  collision resilience and contact-reactive perching,'' {\em Soft Robotics},
  2023.

\bibitem{SP+20}
V.~Serbezov {\em et~al.}, ``Application of multi-axis force/torque sensor
  system,'' {\em IOP Conf. Mater. Sci. and Eng.}, vol.~878, no.~1, p.~012039,
  2020.

\bibitem{OH+18}
A.~Ollero {\em et~al.}, ``The aeroarms project,'' {\em IEEE Robot. Autom. Mag},
  vol.~25, no.~4, pp.~12--23, 2018.

\bibitem{RC+14}
F.~Ruggiero {\em et~al.}, ``Impedance control of vtol uavs with a
  momentum-based external generalized forces estimator,'' in {\em IEEE Int.
  Conf. Robot. Autom.}, pp.~2093--2099, 2014.

\bibitem{RC+15}
F.~Ruggiero {\em et~al.}, ``Passivity-based control of vtol uavs with a
  momentum-based estimator of external wrench and unmodeled dynamics,'' {\em
  Robot. Auto. Sys.}, vol.~72, pp.~139--151, 2015.

\bibitem{YS+14}
B.~Y{\"u}ksel {\em et~al.}, ``A nonlinear force observer for quadrotors and
  application to physical interactive tasks,'' in {\em IEEE/ASME Int. Conf.
  Adv. Intl. Mech.}, pp.~433--440, 2014.

\bibitem{RM+19}
M.~Ryll {\em et~al.}, ``6d interaction control with aerial robots: The flying
  end-effector paradigm,'' {\em Int. J. Robot. Research}, vol.~38, no.~9,
  pp.~1045--1062, 2019.

\bibitem{MS16}
C.~D. McKinnon and A.~P. Schoellig, ``Unscented external force and torque
  estimation for quadrotors,'' in {\em IEEE/RSJ Int. Conf. Intl. Robot. Sys.},
  pp.~5651--5657, 2016.

\bibitem{TOH17}
T.~Tomi{\'c} {\em et~al.}, ``External wrench estimation, collision detection,
  and reflex reaction for flying robots,'' {\em IEEE Trans. Robot.}, vol.~33,
  no.~6, pp.~1467--1482, 2017.

\bibitem{PB+19}
P.~Pf{\"a}ndler {\em et~al.}, ``Flying corrosion inspection robot for corrosion
  monitoring of civil structures--first results,'' in {\em SMAR Conf. on Smart
  Moni. Assess. Rehab. Civil Struc. Prog.}, pp.~We--4, 2019.

\bibitem{HZ22}
J.~Hu, S.~Zhang, E.~Chen, and W.~Li, ``A review on corrosion detection and
  protection of existing reinforced concrete (rc) structures,'' {\em
  Construction and Building Materials}, vol.~325, p.~126718, 2022.

\bibitem{AD16}
K.~Alexis {\em et~al.}, ``Aerial robotic contact-based inspection: planning and
  control,'' {\em Autonomous Robots}, vol.~40, pp.~631--655, 2016.

\bibitem{KB+20}
K.~Bodie {\em et~al.}, ``Active interaction force control for contact-based
  inspection with a fully actuated aerial vehicle,'' {\em IEEE Trans. Robot.},
  vol.~37, no.~3, pp.~709--722, 2020.

\bibitem{TC+19}
M.~Tognon {\em et~al.}, ``A truly-redundant aerial manipulator system with
  application to push-and-slide inspection in industrial plants,'' {\em IEEE
  Robot. Autom. Lett}, vol.~4, no.~2, pp.~1846--1851, 2019.

\bibitem{B99}
B.~Brogliato, {\em Nonsmooth mechanics}, vol.~3.
\newblock Springer, 1999.

\bibitem{ZB10}
J.-C. Zufferey, A.~Beyeler, and D.~Floreano, ``Optic flow to steer and avoid
  collisions in 3d,'' {\em Flying Insects and Robots}, pp.~73--86, 2010.

\bibitem{SB09}
D.~Schafroth, S.~Bouabdallah, C.~Bermes, and R.~Siegwart, ``From the test
  benches to the first prototype of the mufly micro helicopter,'' {\em J.
  Intel. Robot. Sys.}, vol.~54, pp.~245--260, 2009.

\bibitem{HB10}
R.~He, A.~Bachrach, and N.~Roy, ``Efficient planning under uncertainty for a
  target-tracking micro-aerial vehicle,'' in {\em IEEE Int. Conf. Robot.
  Autom.}, pp.~1--8, 2010.

\bibitem{SM11}
S.~Shen, N.~Michael, and V.~Kumar, ``Autonomous multi-floor indoor navigation
  with a computationally constrained mav,'' in {\em IEEE Int. Conf. Robot.
  Autom.}, pp.~20--25, 2011.

\bibitem{SA14}
D.~Scaramuzza {\em et~al.}, ``Vision-controlled micro flying robots,'' {\em
  IEEE Robot. Autom. Mag}, vol.~21, no.~3, pp.~26--40, 2014.

\bibitem{KM19}
N.~Khedekar {\em et~al.}, ``Contact--based navigation path planning for aerial
  robots,'' in {\em Int. Conf. Robot. Autom.}, pp.~4161--4167, IEEE, 2019.

\bibitem{DN21}
P.~De~Petris {\em et~al.}, ``Resilient collision-tolerant navigation in
  confined environments,'' in {\em IEEE Int. Conf. Robot. Autom.},
  pp.~2286--2292, 2021.

\bibitem{ZM21}
J.~Zha and M.~W. Mueller, ``Exploiting collisions for sampling-based
  multicopter motion planning,'' in {\em IEEE Int. Conf. Robot. Autom.},
  pp.~7943--7949, 2021.

\bibitem{SF16}
J.~L. Schonberger and J.-M. Frahm, ``Structure-from-motion revisited,'' in {\em
  Proc. IEEE Conf. Comput Vis. Patt. Recog.}, pp.~4104--4113, 2016.

\bibitem{CV19}
D.~Cattaneo {\em et~al.}, ``Cmrnet: Camera to lidar-map registration,'' in {\em
  IEEE Intel. Transp. Sys. Conf.}, pp.~1283--1289, 2019.

\bibitem{ML20}
Y.~Mulgaonkar {\em et~al.}, ``The tiercel: A novel autonomous micro aerial
  vehicle that can map the environment by flying into obstacles,'' in {\em Int.
  Conf. Robot. Automation}, pp.~7448--7454, IEEE, 2020.

\bibitem{BK13}
A.~Briod {\em et~al.}, ``Contact-based navigation for an autonomous flying
  robot,'' in {\em IEEE/RSJ Int. Conf. Intl. Robot. Sys.}, pp.~3987--3992,
  2013.

\bibitem{HR+14}
K.~M. Hasan {\em et~al.}, ``Path planning algorithm for autonomous vacuum
  cleaner robots,'' in {\em Int. Conf. Info, Elec. Vis.}, pp.~1--6, IEEE, 2014.

\bibitem{ZP18}
Q.-Y. Zhou, J.~Park, and V.~Koltun, ``{Open3D}: {A} modern library for {3D}
  data processing,'' {\em arXiv:1801.09847}, 2018.

\bibitem{AC+19}
A.~Cristofaro {\em et~al.}, ``Time-optimal control for the hybrid double
  integrator with state-driven jumps,'' in {\em IEEE Conf. Decision. Control},
  pp.~6301--6306, 2019.

\end{thebibliography}

\begin{appendices}
\newpage
\section*{Appendix}\label{appendix}

\begin{table*}[!t]
\fontsize{7pt}{7pt}\selectfont
\centering
\caption{Comparison with existing literature for UAV-based tactile applications}
\vspace{-0.1in}
\label{table:literature_comparison}
\renewcommand{\arraystretch}{1.2}
\begin{tabular}{p{2.6cm}p{1.2cm}p{0.9cm}p{3cm}p{3.3cm}p{1cm}p{1.6cm}p{1.1cm}}
\toprule
\textbf{Article} & \multicolumn{2}{c}{\textbf{Type}} & \textbf{Wrench Information} & \textbf{New UAV Primitives} & \textbf{Planning} & \textbf{Mapping} & \textbf{Min. Time} \\ 
& \textbf{Chassis} & \textbf{Extra} &  & & & \textbf{Contours} &\textbf{Braking} \\ 
\midrule
Bredenbeck A. et. al \cite{B+24} & rigid & finger & finger (joint encoders) & end-effector based   & online & no & no \\ 
Alexis, K. et al \cite{AD16} & rigid & none & -- (planning is independent)  & RRT*
and LKH for TSP & offboard & no & no \\ 
Bodie K. et. al \cite{KB+20} & rigid & arm & force-torque sensors & -- (existing vision + tree-lookup) & online & yes & no \\ 
Tognon, M. et al \cite{TC+19} & rigid & arm & -- (planning is independent) & -- (existing control-aware plan) & offboard & yes & no \\ 
Khedekar N. et al. \cite{KM19} & rigid & none & -- (assumes known surface) & flying-cartwheel, sliding & offline & no & no \\
P. De Petris et al. \cite{DN21} & rigid & flap & flex sensors & adaptive acceleration & online & no & no \\
Mulgaonkar Y. et al. \cite{ML20} & rigid & none & onboard IMU sensors & reflected-ray based waypoint & online & 2D rectilinear & no \\
Briod A. et al. \cite{BK13} & deformable & none & onboard IMU sensors & random exploration based wp & online & only walls & no  \\
\textit{\textbf{XPLORER}} & \textit{\textbf{deformable}} & \textit{\textbf{none}} & \textit{\textbf{arm orientation sensor with onboard IMU}} & \textbf{\textit{contact-based reactive primitives}} \textbf{\textit{(tactile turning, traversal, ricocheting)}} & \textit{\textbf{online}} & \textit{\textbf{wall, pipe, box, trashcan}} & \textit{\textbf{yes}} \\ 
\bottomrule
\end{tabular}
\vspace{-0.1in}
\end{table*}
\begin{table*}[!t]
\fontsize{7pt}{7pt}\selectfont
\centering
\caption{Comparison with existing literature for UAV wrench estimation techniques}
\vspace{-0.1in}
\label{table:literature_comparison_wrench}
\renewcommand{\arraystretch}{1.2}
\begin{tabular}{p{2.9cm}p{0.6cm}p{0.75cm}p{2.7cm}p{3.5cm}p{1.4cm}p{3.3cm}}
\toprule
\textbf{Article} & \multicolumn{2}{c}{\textbf{New Method}} & \textbf{Sensors Needed} & \textbf{Estimation Technique} & \textbf{Collision } & \textbf{Application Demonstrated} \\ 
& \textbf{Force} & \textbf{Torque} & & & \textbf{Isolation} & \textbf{}  \\ 
\midrule
Ruggiero, F. et al. \cite{RC+15} & yes & yes & onboard IMU sensors & momentum-based  & no & low-level control design\\
Y{\"u}ksel, B. et al. \cite{YS+14} & yes & yes & -- & a Lyapunov based observer & -- & -- (only simulations) \\
McKinnon \& Schoellig \cite{MS16} & yes & yes & onboard IMU sensors & UKF based observer & no & admittance control design\\
Tomi{\'c}, Teodor \cite{TOH17} & yes & yes & onboard IMU sensors & hybrid-based  & yes & collision-recovery \\
\textit{\textbf{XPLORER}} & \textit{\textbf{yes}} & \textit{\textbf{no}} & \textbf{\textit{onboard IMU, 4 BNO055}} & \textit{\textbf{acceleration-based, hybrid fusing of CoM and proprioceptive}} & \textit{\textbf{collision arm identified}} & \textit{\textbf{tactile-based exploration, mapping and collision recovery}} \\ 
\bottomrule
\end{tabular}
\vspace{-0.2in}
\end{table*}
\subsection{Quadrotor Dynamics}
Considering the net body thrust, $f \in \mathbb{R}$, and the body torques, $\boldsymbol{\tau} \in \mathbb{R}^3$, as the control inputs, the rigid-body dynamics can be written as:
\begin{subequations}\label{eqn:dynamics}
\vspace{-0.1in}
\begin{align}
\centering
\boldsymbol{\dot{x}} &= \boldsymbol{v} \nonumber \\ 
m\boldsymbol{\dot{v}} &= m g \boldsymbol{e}_3 - f \boldsymbol{R} \boldsymbol{e}_3 + \boldsymbol{\delta_{f}} \label{eqn:trans_dynamics} \\  
\boldsymbol{\dot{R}} &= \boldsymbol{R}\hat{ \boldsymbol{\Omega}} \nonumber \\ 
\boldsymbol{H}\dot{ \boldsymbol{\Omega}} &- [\boldsymbol{H} \boldsymbol{\Omega}]_\times  \boldsymbol{\Omega} =  \boldsymbol{\tau} + \boldsymbol{\delta_{\tau}} \label{eqn:rot_dynamics}
\end{align}
\vspace{-.01in}
\end{subequations}
\hspace{-0.09in}where $m$, $\boldsymbol{H}$ denote the mass and inertia of XPLORER respectively, and $\boldsymbol{x} \in \mathbb{R}^3$ and $\boldsymbol{v} \in \mathbb{R}^3$ denote the 3D position and translational velocity, respectively. 
The other symbols adhere to literature \cite{PM+20}. 
The terms $ \boldsymbol{\delta_{f}} \in \mathbb{R}^3$ and $ \boldsymbol{\delta_{\tau}} \in \mathbb{R}^3$ denote the lumped external forces and torques respectively applied on the system. The \textit{hat map} $\hat{\cdot}: \mathbb{R}^3 \xrightarrow{} \mathsf{SO(3)}$ is a symmetric matrix operator defined by the condition that $ \boldsymbol{\hat{x}y = x \times y} ~\forall~  \boldsymbol{x,y } \in \mathbb{R}^3$ and $[.]_\times$ is the skew symmetric cross product matrix.
\vspace{-0.1in}
\subsection{Rotation from the Arm Frame to Body Frame}\label{sec:appx_Rarm}
The rotation matrix for converting a vector in the arm frame $_{a_i}\mathcal{F}$, to body frame $_b\mathcal{F}$, is given by (\ref{eqn:arm_R}) below
\rev{
\begin{equation}\label{eqn:arm_R}
_{a_i}^b\boldsymbol{R} = \begin{bmatrix}
\cos \varphi_i & - \sin \varphi_i & 0\\
\sin \varphi_i & \cos \varphi_i & 0\\
0 & 0 & 1
\end{bmatrix}
\end{equation}
}
\rev{with $\varphi_i$ being the arm angle deflection for the $i^{th}$ arm calculated as $\varphi_i = (\nu_i + \mu_i + \theta_i)$ and
}
\rev{$\nu_i = \frac{-\pi}{2}$ for $i =1,2$ and $\nu_i = \frac{+\pi}{2}$ for $i =3,4$ and $\mu_i$ = $\{\frac{3\pi}{4},\frac{\pi}{4},\frac{-\pi}{4},\frac{-3\pi}{4}\}$ for $i=\{1,2,3,4\}$ respectively. Along with (\ref{eqn:arm_wrench}), this gives the force estimate from spring action in $_w\mathcal{F}$ frame.}

\vspace{-0.1in}
\subsection{Boundedness and Convergence of the Wrench Estimator}\label{sec:appx_estimator_proof}
\textcolor{black}{In this section we comment on the stability of the force estimator which is critical for tactile-primitive selection. We prove that the estimate is bounded at all times and converges to the arm-based estimate as $t \rightarrow \infty$. Let $\boldsymbol{\hat{\delta}_{f_{a}}} \coloneqq \Upsilon \cdot \sum_{i=1}^4  \boldsymbol{\hat{\delta}}_{\boldsymbol{f}_i} $ denote the estimate from the spring-based arm wrench.}

\textcolor{black}{\textit{Assumption 1}: Boundedness of estimated signals $\boldsymbol{\hat{\delta}_{f_{CoM}}}$ and $\boldsymbol{\hat{\delta}_{f_{a}}}$ and their derivatives $\boldsymbol{\dot{\hat{\delta}}_{f_{CoM}}}$, $\boldsymbol{\dot{\hat{\delta}}_{f_{a}}}$. We assume that the individual estimates from the acceleration-based method employed on the UAV and the spring-damper arm are bounded.}

\textcolor{black}{\textit{Remark 1:} Considering that most real-world estimators such as those proposed in \cite{TOH17} are employed for force feedback control, it is reasonable to assume that acceleration-based methods are bounded and provide bounded estimates even in the presence of uncertainties.}

\textcolor{black}{\textit{Assumption 2: } The derivative of $\boldsymbol{\dot{\hat{\delta}}_{f_{CoM}}} \rightarrow 0 $ as $t \rightarrow \infty$.}

\textcolor{black}{\textit{Remark 2:} The acceleration-based estimator dynamics are second-order filter dynamics and hence it is assumed that the signal $\boldsymbol{\dot{\hat{\delta}}_{f_{CoM}}}$ converges to a bounded value implying the rate of change at the convergence is near zero.} 

\textcolor{black}{\textit{Proposition: If $\boldsymbol{\kappa_f}$ is chosen such that $\boldsymbol{\kappa_f} = h(\boldsymbol{\dot{\hat{\delta}}_{f_{CoM}}})$
where $h(\cdot)$ is a function that continuously increases from 0 to 1 as the magnitude of $\boldsymbol{\dot{\hat{\delta}}_{f_{CoM}}}$ decreases, then $\boldsymbol{\kappa_f} \rightarrow 1 $ if $\boldsymbol{\dot{\hat{\delta}}_{f_{CoM}}} \rightarrow 0$ and vice-versa. This allows for the estimate to be bounded at all times.}}

\textcolor{black}{\textit{Proof:} We now discuss the boundedness and convergence (steady state) of the proposed force estimator via a break-up into following three different stages.
\begin{enumerate}
\item \textit{Steady-state phase:} From \textit{Assumption 2}, at steady-state, the $\boldsymbol{\dot{\hat{\delta}}_{f_{CoM}}} \rightarrow 0$ and hence the proposed estimator will converge to the arm-based one, that is $\boldsymbol{\dot{\hat{\delta}}_{f}} \rightarrow \boldsymbol{\dot{\hat{\delta}}_{f_{a}}}$.
\item \textit{Transient phase:} In the transient phase, $\boldsymbol{\dot{\hat{\delta}}_{f_{CoM}}} \rightarrow 1$ and hence the proposed estimator will converge to the CoM-based one, that is $\boldsymbol{\dot{\hat{\delta}}_{f}} \rightarrow \boldsymbol{\dot{\hat{\delta}}_{f_{CoM}}}$.
\item \textit{Transition between transients and steady-state:}  Finally by ensuring that $h(\cdot) \in [0,~1]$ is a smooth function and from \textit{Assumptions 1 and 2}, the force estimate from the proposed estimator is bounded at all times.  
\end{enumerate}}

\vspace{-0.1in}
\subsection{System Identification for Arm Dynamics}
\begin{figure}[t]
\centering
\subfloat[Arm System Identification \label{fig:arm_char}]{\includegraphics[width = 0.24\textwidth, trim = 0cm 0cm 0cm 0cm, clip]{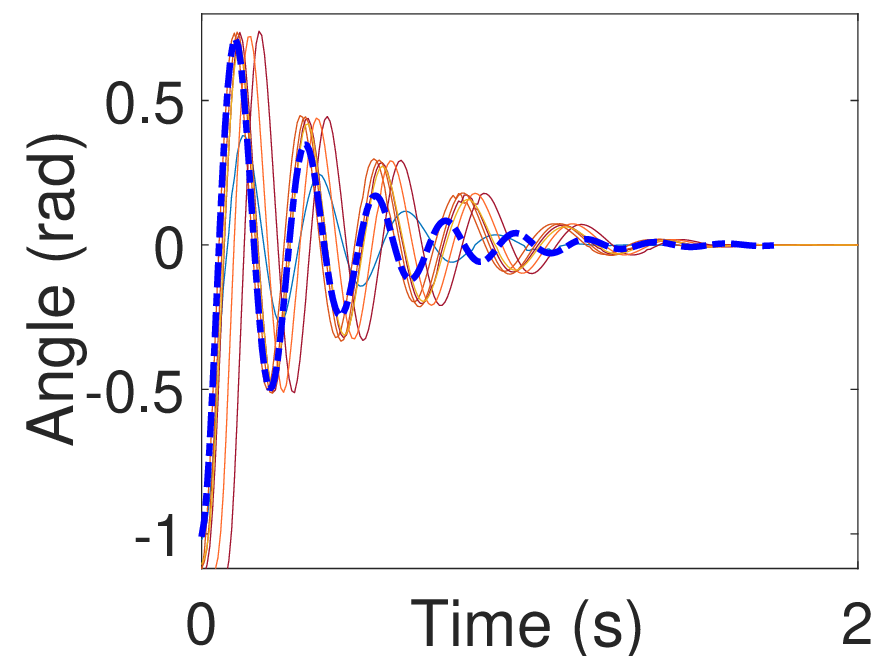}}
\subfloat[Pulley Experiment \label{fig:pulley}]{\includegraphics[width = 0.24\textwidth]{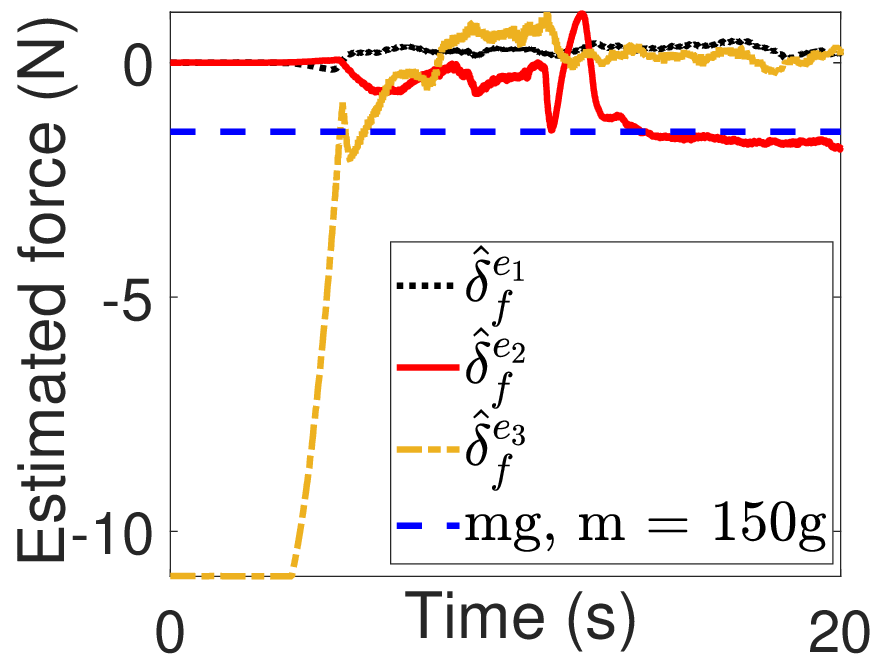}}
\caption{(i) Arm dynamics characterization with a step input. The system identification is done in MATLAB to obtain the damping and spring coefficients. (ii) Validation of the proposed force estimator when there is no contact.}
\vspace{-0.2in}
\end{figure}

For the system identification of the arm dynamics, the inertia about $z$-axis for the arm, $\mathcal{J}_{zz}$ was calculated using SolidWorks to be 0.0015kgm$^2$.  We design an experiment where the base of the XPLORER is held by a clamp and the torsional spring-based arm is first loaded to a certain fixed position using a force sensor. Then it is released and the motion capture system is used to record the trajectory of the arm (shown in Fig. \ref{fig:arm_char}). We then use the MATLAB System ID toolbox to obtain the $b$ and $k$ values as 0.009 and 1.307 for values in Section \ref{sec:arm_wrench}.

\vspace{-0.3in}
\subsection{Additional Results for Force Estimation at No Contact}
In this section we present additional results for the pulley-mass experiment which is to show that when there is no contact, the estimator correctly relies on the CoM-based estimate and the interaction controller successfully uses these estimates to exert the desired force. A mass of 150g was attached to a pulley and directly anchored to the CoM of XPLORER. A desired force equal to weight of the test-mass was commanded. As shown by the results in Fig. \ref{fig:pulley}, the proposed estimator successfully utilizes the CoM-based estimate to perform the task. The results were consistent over three trials for each of 100g, 150g and 200g test masses.

\vspace{-0.1in}
\subsection{Algorithms}\label{sec:appx_algo}
This section presents \textit{Exploration, Tactile turning} and \textit{Tactile-traversal} in an algorithmic format. The position and yaw reference before current time-step is denoted by $[\boldsymbol{x_{sp}^*}(k) ~~\psi^*(k)]^T$ and is used to reshape the reference at current step, $[\boldsymbol{x_{sp}}(k) ~~\psi(k)]^T$. Furthermore, we transform the reference into the inertial frame whenever necessary via 
\begin{equation}
\label{eqn:transform}
[x_{sp}^b ~ y_{sp}^b ~ z_{sp}^b]^T = _b^w\boldsymbol{R} [x_{sp} ~ y_{sp} ~ z_{sp}]^T
\end{equation}
where $[x_{sp}^b ~ y_{sp}^b ~ z_{sp}^b]$ denote the reference in the body frame, $_b\mathcal{F}$, and $[x_{sp} ~ y_{sp} ~ z_{sp}]$ denote them in inertial frame $_w\mathcal{F}$. Similarly, \ref{eqn:transform} is also used to calculate the current position in body frame, $\boldsymbol{x}^b$, from current position in inertial frame, $\boldsymbol{x}$. 

\textit{Note:} $AdmittanceController()$ in Algorithms 1-3 refers to implementation of (\ref{eqn:interaction_controller}) from the main article which reshapes the position and yaw reference based on desired wrench.

\vspace{-0.1in}
\subsection{Time-optimal control with state jumps}
\subsubsection{Problem setup: }Consider the following double integrator dynamics for a continuous-time system with states $\boldsymbol{\tilde{x}} \in \mathbb{R}^2$:
\begin{equation}\label{eqn:single_int}
\begin{aligned}
\dot{\tilde{x}}_1 = \tilde{x}_2, 
~~~\dot{\tilde{x}}_2 = u
\end{aligned}
\end{equation}
It is assumed that the jump pattern is induced by the flow set $\mathcal{C}:= \{\boldsymbol{\tilde{x}}\in \mathbb{R}^2\}$ and the jump set which denotes the locations for collision is denoted by
$ \mathcal{D} = \{\boldsymbol{\tilde{x}}\in \mathbb{R}^2 | \tilde{x}_1 = a, \tilde{x}_2 \leq 0 \} $. Here, $a$ is the location of objects that XPLORER can collide on. Furthermore, let the hybrid system be described by:
\begin{equation}\label{eqn:hybrid_dynamics}
\begin{aligned}
\boldsymbol{\dot{\tilde{x}}} = \boldsymbol{A}\boldsymbol{\tilde{x}} +\boldsymbol{B}u \text{ if } \boldsymbol{\tilde{x}} \in \mathcal{C}, 
\boldsymbol{\tilde{x}^+} = \boldsymbol{E}\boldsymbol{\tilde{x}} \text{ if } \boldsymbol{\tilde{x}} \in \mathcal{D}
\end{aligned}
\end{equation}
where $\boldsymbol{A} \in \mathbb{R}^{2\times2}$ is state matrix, $\boldsymbol{B} \in \mathbb{R}^{2\times1}$ is input matrix and $\boldsymbol{E} \in \mathbb{R}^{2\times2}$ denotes the state evolution during the jump. 

The conventional time-optimal control for a system to reach $(0,0)$ from a state $\boldsymbol{\tilde{x}}_0$
is given by the \textit{bang-bang} control law where the control only takes the values $\{ -1,1\}$ and at most one switch is required. More specifically, the trajectories are characterized by the family of the parabolas: 
\begin{equation*}
\medmath{
\begin{aligned}
\begin{cases}
u = +1 \\
\begin{cases}
	\tilde{x}_1(t) = +\frac{1}{2}t^2 + \alpha t + \beta \\
	\tilde{x}_2(t) = +t + \alpha
\end{cases} \\
\tilde{x}_1 = \frac{1}{2}\tilde{x}_2^2 + \beta - \frac{\alpha^2}{2}
\end{cases}
\begin{cases}
u = -1 \\
\begin{cases}
	\tilde{x}_1(t) = -\frac{1}{2}t^2 + \alpha t + \beta \\
	\tilde{x}_2(t) = -t + \alpha
\end{cases} \\
\tilde{x}_1 = -\frac{1}{2}\tilde{x}_2^2 + \beta - \frac{\alpha^2}{2}
\end{cases}
\end{aligned}}
\end{equation*}
\begin{algorithm}[t]
\footnotesize
\caption{Exploration}\label{alg:Exploration}

\SetKwInput{KwInput}{Input}                
\SetKwInOut{Parameter}{Parameters}         
\SetKwInput{KwOutput}{Output}              
\DontPrintSemicolon
\KwInput{$ \boldsymbol{x^b}, [\boldsymbol{x_{sp}^*}~\psi_{sp}^*]^T, \dot{\psi} \hspace{0.1 cm} \& \hspace{0.1 cm} {\boldsymbol{\hat{\delta}_f^b}} $}
\KwOutput{$ [\boldsymbol{x_{sp}^b}~\psi_{sp}]^T $ \& $\Gamma$ }
\Parameter{$d_{step}$}

\If{$ |\dot{\psi}| < \dot{\psi}_{o} \hspace{0.1 cm} \& \hspace{0.1 cm} |{\boldsymbol{\hat{\delta}_f^b}}| < {{{\delta}}}_{0} $}
{
$ x_{sp}^b(k) = x^b (k) + d_{step} $
\newline
$ y_{sp}^b(k) = (y_{sp}^{*})^b (k)$
\newline
$ \psi_{sp}(k) = \psi_{sp}^*(k) $
}
\ElseIf{$ |\dot{\psi}| > \dot{\psi}_{o} $}
{   
$\Gamma = 2 $
}
\Else
{   
$\Gamma = 3 $
}
\vspace{-0.2in}
\end{algorithm}
\begin{algorithm}[t]
\footnotesize
\caption{Tactile Turning}\label{alg:Tactile_Turning}

\SetKwInput{KwInput}{Input}                
\SetKwInOut{Parameter}{Parameters}         
\SetKwInput{KwOutput}{Output}              
\DontPrintSemicolon
\KwInput{$ [\boldsymbol{x_{sp}^*}~\psi_{sp}^*]^T, \dot{\psi} \hspace{0.1 cm} \& \hspace{0.1 cm} {\boldsymbol{\hat{\delta}_f^b}}$}
\KwOutput{$ [{\boldsymbol{x_{sp}}}~\psi_{sp}]^T $ \& $\Gamma$}
\Parameter{$\mathcal{W}, dt$}

\If{$|\dot{\psi}| > \dot{\psi}_{0}$ \& $|{\boldsymbol{\hat{\delta}_f^b}}| < {{{\delta_{\psi}}}}_{0}$}
{$\boldsymbol{x_{sp}}(k) = \boldsymbol{x_{sp}^*}(k)$
\newline
$\dot{\psi}_{sp}(k) = sgn(\dot{\psi})\mathcal{W} \cdot dt$
}
\Else
{
$\Gamma = 3$
}
\vspace{-0.3in}
\end{algorithm}
\begin{algorithm}
\footnotesize
\caption{Tactile Traversal}\label{alg:Tactile_Traversal}

\SetKwInput{KwInput}{Input}                
\SetKwInOut{Parameter}{Parameters}         
\SetKwInput{KwOutput}{Output}              
\SetKwInput{CollisionNormals}{State Update}
\SetKwInput{MoveDirection}{State Update}
\SetKwInput{TrajGenerate}{Trajectory Generation}
\SetKwInput{Main}{Main}
\DontPrintSemicolon

\KwInput{$ \boldsymbol{x^b}, \dot{\psi}, \hspace{0.1 cm} [\boldsymbol{x^*_{sp}} ~\psi_{sp}^*]^T \hspace{0.1 cm} \& \hspace{0.1 cm} {\boldsymbol{\hat{\delta}_f^b}} \hspace{0.1cm} \& \hspace{0.1cm} \hat{\delta}_{\tau}^{e_3} $}
\KwOutput{$ [\boldsymbol{x_{sp}^b}~\psi_{sp}]^T $ \& $\Gamma$ }
\Parameter{$\delta_0, \boldsymbol{\delta_{f}}_{des}$}

\SetKwFunction{CollisionNormal}{CollisionNormal}
\SetKwFunction{MoveDirection}{MoveDirection}
\SetKwFunction{TrajectoryGenerate}{TrajectoryGeneration}
\SetKwFunction{Main}{Main}    
\SetKwProg{Fn}{Function}{:}{}
\Fn{\CollisionNormal{}}{
\If{$ \lambda = ``+X" or ``-X" $}
{
\If{${{{\hat{\delta}_f}^{b_1}}} > {{{{\delta}}}_{0}} $ }
{
$\boldsymbol{C_n} = [1,0,0,0] $
}
\ElseIf{${{{\hat{\delta}_f}^{b_1}}} < -{{{{\delta}}}_{0}} $}
{
$\boldsymbol{C_n} = [0,1,0,0] $
}
}
\ElseIf{$ \lambda = ``+Y" or ``-Y" $}
{
\If{${{{\hat{\delta}_f}^{b_2}}} > {{{{\delta}}}_{0}}$ }
{
$\boldsymbol{C_n} = [0,0,1,0] $
}
\ElseIf{${{{\hat{\delta}_f}^{b_2}}} < -{{{{\delta}}}_{0}} $}
{
$\boldsymbol{C_n} = [0,0,0,1] $
}
}

\KwRet $\boldsymbol{C_n}$
}

\SetKwProg{Fn}{Function}{:}{}
\Fn{\MoveDirection{}}{
$\boldsymbol{C_n} \leftarrow $ ContactNormal()
\newline
\If{$\boldsymbol{C_n} == [1,0,0,0] $} {$ \lambda = ``+Y" $}
\ElseIf{$\boldsymbol{C_n} == [0,1,0,0] $} {$ \lambda = ``-Y" $}
\ElseIf{$\boldsymbol{C_n} == [0,0,1,0] $} {$ \lambda = ``-X" $}
\ElseIf{$\boldsymbol{C_n} == [0,0,0,1] $} {$ \lambda = ``+X" $}

\KwRet $\lambda$
}

\SetKwProg{Fn}{Function}{:}{}
\Fn{\TrajectoryGenerate{}}{

${\lambda} \leftarrow $ MoveDirection()        \newline
$ \psi_{sp} = \psi_{sp}^* $
\newline
\If{$ \lambda = ``+X" $}
{
$ x_{sp}^b(k) = x^b + d_{step} $
\newline
$\boldsymbol{\delta_{f}}_{des} = [0~\Delta_{f_{des}}~0]^T, \delta_{\tau_{des}} = \hat{\delta}_{\tau}^{e_3}$
\newline
$ \boldsymbol{x_{sp}}(k) = AdmittanceController(\boldsymbol{\delta_{f}}_{des},\delta_{\tau_{des}})$

}
\ElseIf{$ \lambda = ``-X" $}
{
$ x_{sp}^b(k) = x^b - d_{step} $
\newline
$\boldsymbol{\delta_{f}}_{des} = [0~-\Delta_{f_{des}}~0]^T, \delta_{\tau_{des}} = \hat{\delta}_{\tau}^{e_3}$ 
\newline
$ \boldsymbol{x_{sp}}(k) = AdmittanceController(\boldsymbol{\delta_{f}}_{des},\delta_{\tau_{des}})$

}
\ElseIf{$ \lambda = ``+Y" $}
{
$ y_{sp}^b(k) = y^b + d_{step} $
\newline
$\boldsymbol{\delta_{f}}_{des} = [\Delta_{f_{des}}~0~0]^T, \delta_{\tau_{des}} = \hat{\delta}_{\tau}^{e_3}$
\newline
$ \boldsymbol{x_{sp}}(k) = AdmittanceController(\boldsymbol{\delta_{f}}_{des},\delta_{\tau_{des}})$

}
\ElseIf{$ \lambda = ``-Y" $}
{
$ y_{sp}^b(k) = y^b - d_{step} $
\newline
$\boldsymbol{\delta_{f}}_{des} = [-\Delta_{f_{des}}~0~0]^T, \delta_{\tau_{des}} = \hat{\delta}_{\tau}^{e_3}$
\newline
$ \boldsymbol{x_{sp}}(k) = AdmittanceController(\boldsymbol{\delta_{f}}_{des},\delta_{\tau_{des}})$
}
}
\rev{
\SetKwProg{Fn}{Function}{:}{}
\Fn{\Main{}}{
~~CollisionNormal()
\newline
MoveDirection()
\newline
TrajectoryGeneration()
}
}      
\vspace{-0.3in}
\end{algorithm}
with the switching curve
$ \Pi:= \{ \boldsymbol{\tilde{x}} : \tilde{x}_1 + \frac{1}{2}\tilde{x}_2|\tilde{x}_2| \} $ and parameters $\alpha, \beta$.
Furthermore, given the initial condition, the optimal time is given by:
\begin{equation}\label{eqn:0jumpt0}
\medmath{
t_0^*({\tilde{x}_0}) = \begin{cases}
2\sqrt{\frac{1}{2}\tilde{x}_{2_0}^2 + \tilde{x}_{1_0} }+\tilde{x}_{2_0}, {\tilde{x}_0} \in S^- \\
| \tilde{x}_{2_0} |, {\tilde{x}_0} \in \Pi \\
2\sqrt{\frac{1}{2}\tilde{x}_{2_0}^2 - \tilde{x}_{1_0} }-\tilde{x}_{2_0}, {\tilde{x}_0} \in S^+
\end{cases}}
\end{equation}
where $\mathbb{R}^2 = S^- \cup \Pi \cap S^+ $, $S^{\pm}$ stands for the regions below and above the switching curve $\Pi$.

\subsubsection{Time optimal control with 1 state jump due to collisions}
We now extend the analysis to study the case scenarios where collisions induce a state jump in velocity. For this case, let the jump map $\boldsymbol{E}$ be given by 
\begin{equation}\label{eqn:E}
\medmath{
\boldsymbol{E} = \begin{bmatrix}
1 & 0 \\
0 & e 
\end{bmatrix}}
\end{equation}
\begin{algorithm}[t]
\footnotesize
\caption{Mapping Framework}\label{alg:Mapping}

\SetKwInput{KwInput}{Input}                
\SetKwInOut{Parameter}{Parameters}         
\SetKwInput{KwOutput}{Output}              
\DontPrintSemicolon

\KwInput{$ [x,y,z]^T \hspace{0.1 cm} , \hspace{0.1 cm} {\psi} \hspace{0.1 cm} , \hspace{0.1 cm} \hat{\boldsymbol{\delta}}_f^b \hspace{0.1 cm} , \hspace{0.1 cm} \boldsymbol{C_n} \hspace{0.1 cm} \& \hspace{0.1 cm} \lambda $}
\KwOutput{$ Point \hspace{0.1 cm} Cloud \hspace{0.1 cm} Data $}
\Parameter{$\delta_{map}$}

\If{$|{\boldsymbol{\hat{\delta}_f^b}}| \ge {\delta}_{map} $}
{
\If{$ \lambda (k-1) \hspace{0.1 cm} \neq \hspace{0.1 cm} \lambda (k)$ }
{
Add Corner Block to Point Cloud
}
\Else
{
Add Obstacle to Point Cloud in $\boldsymbol{C_n}$ axis 
}
}
\Else
{
Store Point Cloud Data
}
\vspace{-0.3in}
\end{algorithm}
where $e$ stands for the coefficient of post collision velocity as a function of the coefficient of restitution. This value $e$ can be obtained using a linearization of the collision model in the  Section \ref{sec:NN}. Then any generic point $\tilde{z}$ on the jump set $\mathcal{D}$ can be denoted by $\tilde{z}=(a,\zeta)$ where $a$ denotes the position of the wall and $\zeta$ denotes the pre-collision velocity. The optimal path to $\tilde{z}$ from any given initial point $\tilde{x}_0$ are given by the two branches of parabolas with a switching curve passing through $\tilde{z}$ as $\Phi(\zeta) = \{ \boldsymbol{\tilde{x}}: \frac{1}{2}(\tilde{x}_2 + \zeta)|\tilde{x}_2 - \zeta | - a + \tilde{x}_1\} $.

We can again decompose the state space as $\mathbb{R}^2 = \Psi^- \cup \Phi \cap \Psi^+$ where $\Psi^{\pm}$ is similarly defined as the regions below and above the switching curve $\Phi$. The optimal time taken to reach $\tilde{z}$ from $\tilde{x}_0$ can then be written as:
\begin{equation}\label{eqn:tau0x0}
\medmath{
\begin{aligned}
\tau(\tilde{x}_0,\zeta) = 
\begin{cases}
\Bigg|\zeta + \sqrt{\frac{\zeta^2}{2} + \frac{\tilde{x}_{2_0}^2}{2} - a + \tilde{x}_{1_0}}\Bigg| + \tilde{x}_{2_0} \\
+ \sqrt{\frac{\zeta^2}{2} + \frac{\tilde{x}_{2_0}^2}{2} - a + \tilde{x}_{1_0}}, & \tilde{x}_0 \in \Psi^+(\zeta), \\
\Bigg| \tilde{x}_{2_0} - \zeta\Bigg|, & \tilde{x}_0 \in \Phi(\zeta), \\
\Bigg|\zeta - \sqrt{\frac{\zeta^2}{2} + \frac{\tilde{x}_{2_0}^2}{2} + a - \tilde{x}_{1_0}}\Bigg| - \tilde{x}_{2_0} \\
+ \sqrt{\frac{\zeta^2}{2} + \frac{\tilde{x}_{2_0}^2}{2} + a - \tilde{x}_{1_0}}, & \tilde{x}_0 \in \Psi^-(\zeta)
\end{cases}
\end{aligned}}
\end{equation}
The corresponding optimal control law is given as:
\begin{equation}\label{eqn:u1star}
\medmath{
\begin{aligned}
u_1^* = \begin{cases}
1, & \text{if}~ j = 0, \gamma_0(\tilde{x}(t,0)) < 0 \\
-1, & \text{if}~ j = 0, \gamma_0(\tilde{x}(t,0)) > 0 \\
-sgn(\tilde{x}_2(t,0)-z_2^*(\tilde{x}_0)), & \text{if}~ j = 0, \gamma_0(\tilde{x}(t,0)) = 0 \\
1, & \text{if}~ j = 1, \gamma_1(\tilde{x}(t,1)) < 0 \\
-1, & \text{if}~ j = 1, \gamma_1(\tilde{x}(t,1)) > 0 \\
-sgn(\tilde{x}_2(t,1)), & \text{if}~ j = 1, \gamma_1(\tilde{x}(t,1)) = 0 
\end{cases}
\end{aligned}}
\end{equation}
where the functions $\gamma_0:\mathbb{R}^2 \rightarrow \mathbb{R}$ and $\gamma_1:\mathbb{R}^2 \rightarrow \mathbb{R}$ are:
\begin{equation*}
\medmath{
\begin{aligned}
\gamma_0(\tilde{x}) &= \frac{1}{2}(\tilde{x}_2 + z_2^*(\tilde{x}_0))|\tilde{x}_2 - z_2^*(\tilde{x}_0)|-z_1^*(\tilde{x}_0) + \tilde{x}_1 \\
\gamma_1(\tilde{x}) &= \tilde{x}_1 + \frac{1}{2}\tilde{x}_2|\tilde{x}_2|
\end{aligned}}
\end{equation*}

Furthermore, the 1-jump minimum time is given by
\begin{equation}\label{eqn:t1starx0}
t_1^*(\tilde{x}_0) = \min_u \{ \tau(\tilde{x}_0,\zeta) + t_0^*(\boldsymbol{E}\tilde{z})\}
\end{equation}
where $\boldsymbol{E}\tilde{z}$ denotes the post collision state. The proof follows \cite{AC+19}. We use this result to motivate and introduce \textit{Ricocheting}.

\textit{Remark 3}: For our experiments, we only select ricocheting if the collision-based travel time—calculated as the time to reach the collision node at maximum velocity plus the time from the post-collision velocity to the goal—is shorter than a direct flight under P-PID control. We begin by choosing the maximum velocity XPLORER can attain (from a discrete set) in the closed flight space and verify via the CollisionNet that a reduced rebound state is indeed generated post-collision. Intuitively, $\zeta$ is both upper- and lower-bounded: it must not be so large that traveling to the collision node inflates total maneuvering time, nor so small that it prematurely diverts the UAV from its course. We will derive these bounds formally in our future work.

\begin{figure}
\centering
\includegraphics[trim = 0 1cm 0 0.5cm, clip, width = 0.48\textwidth]{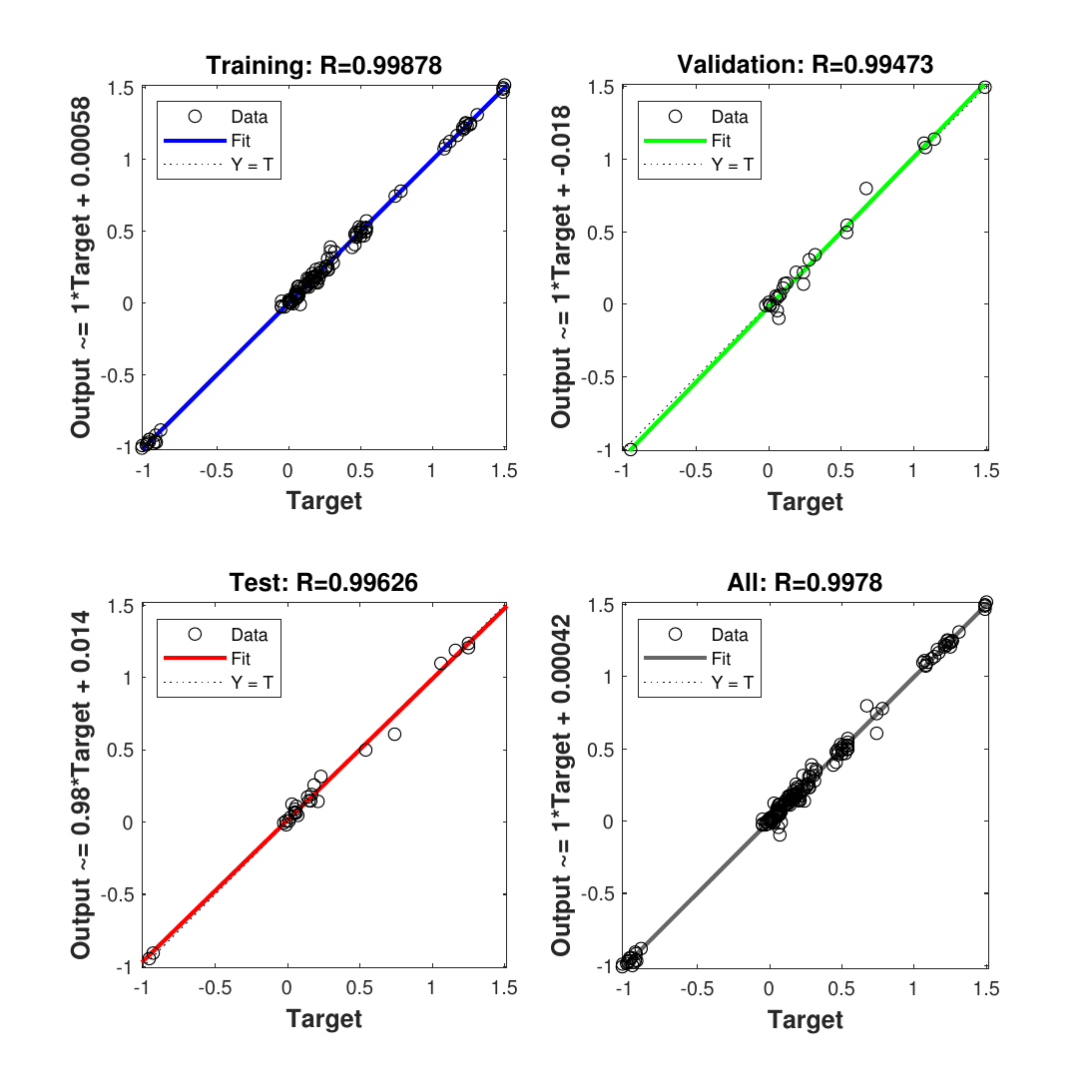}
\caption{Validation results for the collision neural network showing the $R^2$ values obtained for the trained model. We see that the trained network is able to predict the post collision state with high accuracy. }
\label{fig:NN_validation}
\vspace{-0.25in}
\end{figure}

\subsubsection{Illustrative example for time-optimal state jumps}
In this subsection we present an example to motivate \textit{Ricocheting}. For this case, we assume the coefficient of restitution $e = -0.6$ in (\ref{eqn:E}) and a wall at $a = 0.5$. The UAV starts from an initial state $\tilde{x}_0 = [-1~2]^T$, marked by a diamond in Fig. \ref{fig:collision_NN} right. 

For a double integrator system, the right-open parabolic curves represent trajectories under maximum control, $\tilde{u} =  1$ while the left-open curves correspond to minimum control $\tilde{u} =  -1$. The bang-bang control law (without collisions) achieves minimum-time travel in 4 seconds, following the red-magenta trajectory (1-4-5) in Fig. \ref{fig:collision_NN} right, where deceleration continues until the trajectory intersects the switching curve (magenta), followed by acceleration to reach (0,0).

For the collision-inclusive path, a state $(a,\zeta)$ after collision translates to $(a, -e\zeta)$. Using (\ref{eqn:tau0x0}), the minimum time to reach (0,0) via collision and maximum deceleration is 1.756 seconds, with $\zeta^* = 1.667$ and $e = 0.6$. The corresponding bang-bang-jump-bang trajectory is shown by the path 1-2-3-4-5, blue-magenta curved lines, in Fig. \ref{fig:collision_NN} right.

In summary for lower values of $e$, as is the case for deformable quadrotors, \textit{Ricocheting} off an edge near the stopping node will produce a minimum time path for efficient braking.


\vspace{-0.1in}
\subsection{CollisionNet- A Neural Network for post-collision state}\label{sec:NN}
From the set of 60 experiments, a data-driven collision model was developed to study the effects of collisions at various intensities and various angle of incidence for XPLORER. The experimental data collected are plotted as the pre-collision and post-collision velocities in Fig. \ref{fig:collision_NN} left of main article. This collision model is crucial to identify the coefficient of restitution and predict the post-collision state of the vehicle. Since the collisions for the deformable vehicle are highly nonlinear, a nonlinear regression model using neural networks was employed to develop the collision model- CollisionNet. MATLAB Deep Learning Toolbox (MATLAB R2021b) was employed for the training using Levenberg-Marquardt method. Five features were identified to train the model: the $v_x^-,v_y^-$ representing pre-collision state, angle of incidence to the collision plane, and type of collision. The angle of incidence is a function of the orientation of the vehicle, while the type of collision is either collide-to-stop (inelastic collision) or collide-to-decelerate (elastic collision). We choose 75$\%$ of the dataset for training and 25$\%$ for validation. The results for the CollisionNet are given in Fig. \ref{fig:NN_validation} and we see that the accuracy is very high with a $R^2$ value of 0.99. 

It is noticed from Fig. \ref{fig:collision_NN} that for different angles and same pre-collision state, the post collision state is always very similar for the 2D case. This shows that for this particular quadrotor design, angle of collision is not a critical parameter to model and hence justifies our assumption to use a point-mass based double integrator model to ricochet.

\vspace{-0.1in}
\subsection{Additional results for rigid quadrotor}
In this section we present additional results for exploration experiments with a rigid quadrotor of the same dimensions by deploying the exploration scheme developed for XPLORER. We conducted both wall-traversal and the box-traversal experiments out of which only the wall-traversal succeeded.

\begin{figure}[t]
\centering
\subfloat[Wall traversal \label{fig:Traversal_Graph_Rigid_Wall}]{\includegraphics[width = 0.45\textwidth]{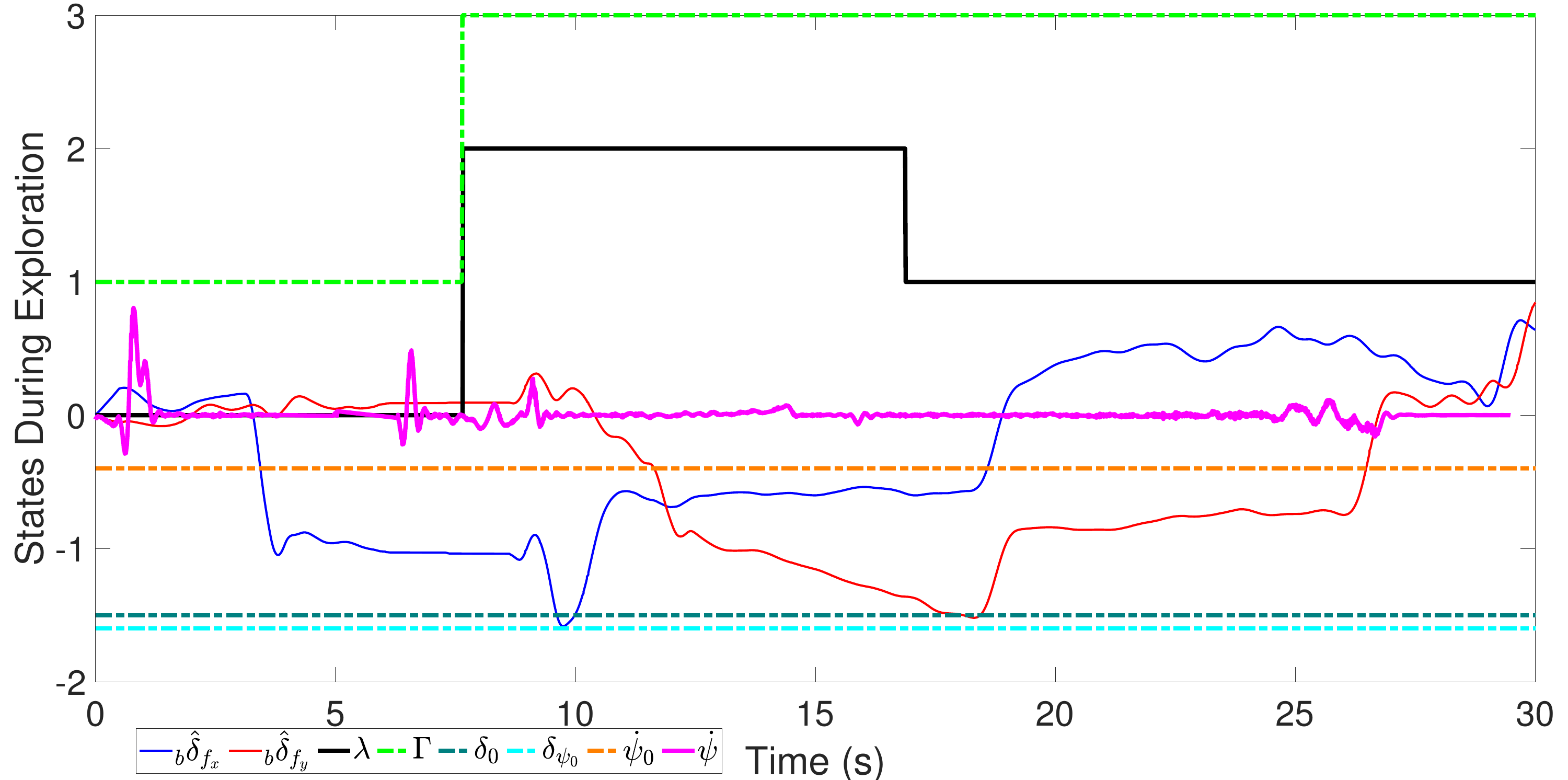}}\\
\vspace{-0.1in}
\subfloat[Failure Case \label{fig:Traversal_Graph_Rigid_Falsetrigger}]{\includegraphics[width = 0.45\textwidth]{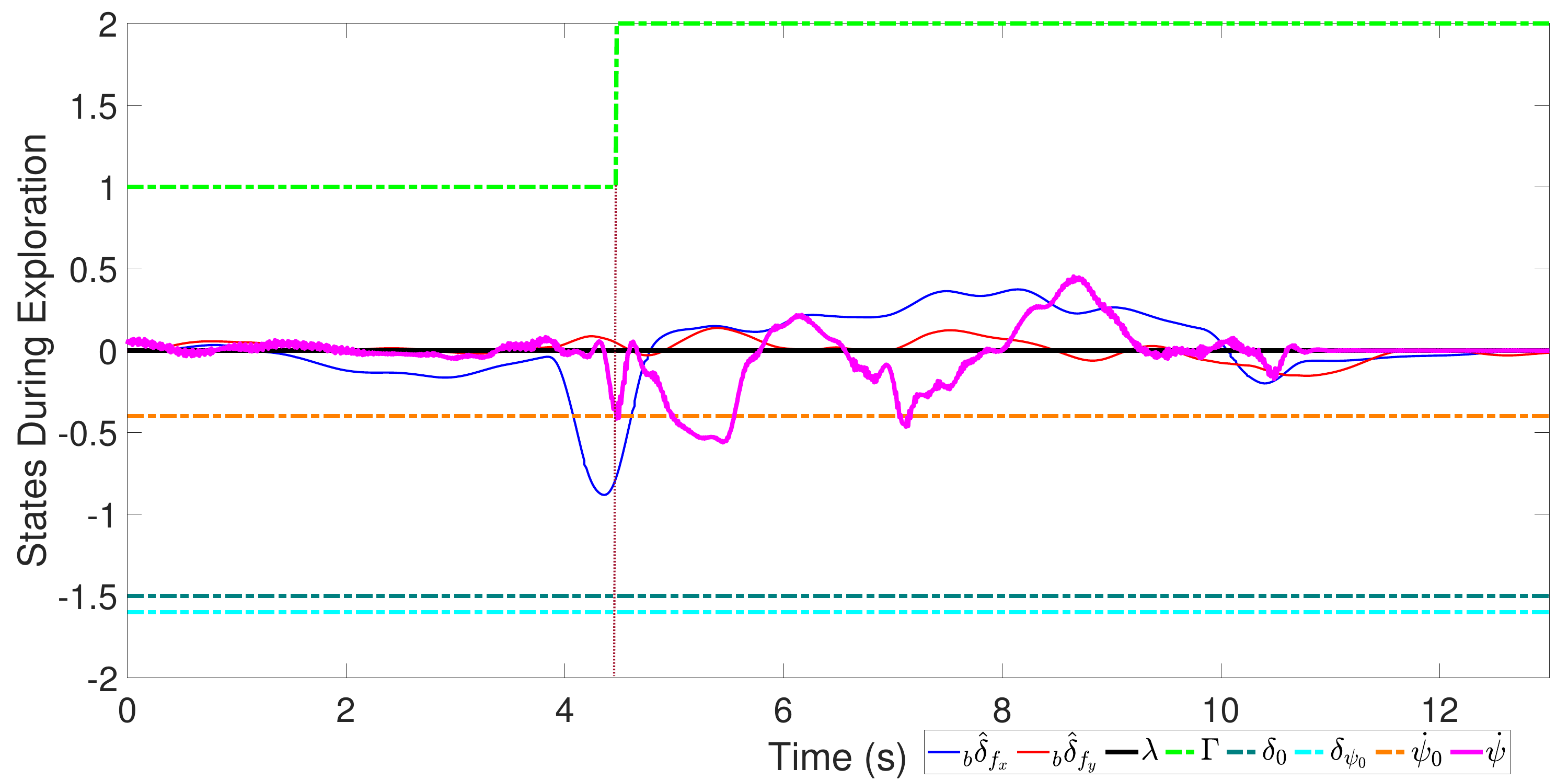}}
\caption{Exploration results for rigid UAV. (a) For a wall-like structure, the performance is as desired, and $\Gamma$ = 1, 3 are engaged.  (b) failures case when the yaw-torque at initial contact is high leading to engaging the \textit{Tactile-turning} state with $\Gamma = 2$ and failing to start the exploration scheme with  $\Gamma = 3$ that is \textit{Tactile-traversal.}}
\label{fig:rigid_exploration_results}
\vspace{-0.2in}
\end{figure}

\rev{For the wall traversal results shown in Fig. \ref{fig:Traversal_Graph_Rigid_Wall}, the rigid UAV takes off and upon making initial contact, applies the desired force on the wall and initiates the \textit{Tactile-traversal}. When it detects the second wall, it engages the traversal motion in the body $+Y$ direction as shown by the black line.}


\rev{Finally, another unpredictable behavior was observed with the rigid UAV, as shown in Fig. \ref{fig:Traversal_Graph_Rigid_Falsetrigger}. Sometimes when the rigid UAV makes impact, due to the high rebound velocities, and the presence of yaw admittance control given by (\ref{eqn:interaction_controller}), there is a high yaw-rate which falsely triggers the \textit{Tactile-turning} and the entire exploration fails. However this behavior is not observed with the XPLORER due to the damping effect and shock-absorbing nature of the torsion spring in the arm.}


\end{appendices}

\end{document}